\documentclass{article}

    \PassOptionsToPackage{numbers}{natbib}

     \usepackage[final]{neurips_2019}

\usepackage[utf8]{inputenc} 
\usepackage[T1]{fontenc}    
\usepackage{hyperref}       
\usepackage{url}            
\usepackage{booktabs}       
\usepackage{amsfonts}       
\usepackage{nicefrac}       
\usepackage{microtype}      

\usepackage{amsmath}
\usepackage{amssymb}
\usepackage{amsthm}
\usepackage{bm}
\usepackage{upgreek}
\usepackage{algorithm2e}
\usepackage{todonotes}
\usepackage[inline]{enumitem}
\usepackage{subcaption}

\newcommand{\btheta}{\bm{\theta}}
\newcommand{\bp}{\bm{p}}

\newcommand{\bx}{\bm{x}}
\newcommand{\by}{\bm{y}}
\newcommand{\bz}{\bm{z}}
\newcommand{\bv}{\bm{v}}

\newcommand{\argmax}{\operatorname{argmax}}
\newcommand{\argmin}{\operatorname{argmin}}
\newcommand{\trace}{\operatorname{tr}}
\newcommand{\cov}{\operatorname{Cov}}
\newcommand{\var}{\operatorname{Var}}
\newcommand{\vectorify}{\operatorname{vec}}
\newcommand{\diag}{\operatorname{diag}}
\newcommand{\expect}{\mathbb{E}}
\newcommand{\entropy}{\mathrm{H}}

\newcommand{\goal}{\tau}
\newcommand{\util}{\pi}
\newcommand{\fisherinfo}{\mathbf{I}}

\newcommand{\goral}{\textsc{Goral}}
\newcommand{\synth}{\texttt{synth2}}
\newcommand{\letter}{\texttt{letter}}
\newcommand{\rtpolar}{\texttt{rt-polarity}}

\newtheorem{remark}{Remark}

\newsavebox{\measurebox}

\title{Understanding Goal-Oriented Active Learning\\via Influence Functions}

\author{%
  Minjie Xu \\
  Bloomberg L.P. \\
  London, UK \\
  \texttt{mxu161@bloomberg.net} \\
  \And
  Gary Kazantsev \\
  Bloomberg L.P. \\
  New York, US \\
  \texttt{gkazantsev@bloomberg.net} \\
}

\begin{document}

\maketitle

\begin{abstract}
  Active learning (AL) concerns itself with learning a model from as few labelled data as possible through actively and iteratively querying an oracle with selected unlabelled samples. In this paper, we focus on analyzing a popular type of AL in which the utility of a sample is measured by a specified goal achieved by the retrained model after accounting for the sample's marginal influence. Such AL strategies attract a lot of attention thanks to their intuitive motivations, yet they also suffer from impractically high computational costs due to their need for many iterations of model retraining. With the help of influence functions, we present an effective approximation that bypasses model retraining altogether, and propose a general efficient implementation that makes such AL strategies applicable in practice, both in the serial and the more challenging batch-mode setting. Additionally, we present both theoretical and empirical findings which call into question a few common practices and beliefs about such AL strategies.
\end{abstract}

\section{Introduction}
Active learning (AL)~\citep{settles09alls} allows a model to actively query an oracle for labels with its chosen unlabelled samples, effectively assembling a growing labelled dataset on the fly along with model training. AL has been studied extensively and a large suite of AL strategies have been proposed to date. However, their success in practice has not been entirely consistent~\citep{lowell19podal}. In fact, even research investigating this problem sometimes produces seemingly contradictory findings, e.g. with \citep{evans13wdalw} claiming AL works better under model ``mis-match'' while \citep{mussmannL18deeus} claiming otherwise. In quest of a better common understanding of several popular AL strategies that can all be abstracted as selecting samples to boost an explicitly specified goal on the model, this work leverages \emph{influence functions}~\citep{koh17ubpif} to closely analyse such AL strategies and offers several interesting insights.

We summarize our main contributions as follows,
\begin{enumerate*}[label={\alph*)},font={\bfseries}]
\item Formalizing a general \emph{goal-oriented} AL framework which generalizes many existing AL strategies;
\item Being the first to apply influence functions to the AL setting, significantly reducing its computational cost (especially for the batch-mode setting);
\item Showing both analytically and empirically that using the current model prediction to resolve the unknown label in such AL strategies (which is a common practice) may not be a sensible choice;
\item Demonstrating the difficulties of finding an effective and sensible goal in goal-oriented AL.
\end{enumerate*}

\section{Goal-Oriented Active Learning (\goral{})}
We focus on \emph{pool-based} active learning~\citep{lewis94pool} for classification problems in this paper. An unlabelled data sample is denoted by $\bx\in\mathcal{X}$ and, when it is labelled, $\bz=(\bx,y)$ where $y\in\{1,\dots,K\}$. We then assume there is a large pool of unlabelled samples $\mathcal{U}_{\text{pool}}=\{\bx\}$ from which an AL \emph{strategy} will pick samples to be labelled by an oracle and then to be added into a growing labelled dataset $\mathcal{L}_{\text{train}}=\{\bz\}$ for model training. We restrict ourselves to discriminative probabilistic models $P_{\btheta}(y|\bx)$ and further assume access to an initial labelled dataset $\mathcal{L}_{\text{init}}$ from which we will train an initial model to kick off the active learning process, as well as a (labelled) test set $\mathcal{L}_{\text{test}}$ which we use to measure model performance and accordingly \emph{data efficiency} (i.e. the minimum $|\mathcal{L}_{\text{train}}|$ needed to reach a certain level of model performance), a key metric for comparing various AL strategies. These three datasets $(\mathcal{U}_{\text{pool}},\mathcal{L}_{\text{init}},\mathcal{L}_{\text{test}})$ jointly define the data dependency of an AL instance in our study.

Many AL strategies (e.g.~\citep{lewis94pool,roy01toal,schein07al4lr}) are dictated by a \emph{utility} function $\util(\bx;\hat{\btheta})$ which, based on the current model $\hat{\btheta}$, assigns a utility score to unlabelled samples. In this paper, we focus on a specific type of utility that depends on $\hat{\btheta}$ only through an explicitly defined \emph{goal function} $\goal(\btheta)$ used for evaluating the models (the higher the better, e.g. model accuracy). And we define this \emph{goal-oriented utility} to be the difference in goals before and after accounting for an additional sample $\bx$, i.e.
\begin{align}
\label{Eq:util_goal}
\util_{\text{goal}}(\bx;\hat{\btheta})\triangleq\ell_y\big[\goal\big(\hat{\btheta}_{(\bx,y)}\big)\big]-\goal(\hat{\btheta})\text{,}
\end{align}
where $\hat{\btheta}_{(\bx,y)}$ represents the model obtained from $\mathcal{L}_{\text{train}}\cup\{(\bx,y)\}$, that is the current training set augmented with one additional sample $\bx$ (hypothetically) labelled as $y$, and $\ell_y[\cdot]$ represents an operator that resolves its operand's dependency on $y$ (e.g. $\min_y$, $\expect_y$, etc.). Going forward we call all such AL strategies \emph{\textbf{G}oal-\textbf{Or}iented \textbf{A}ctive \textbf{L}earning} (\goral{}).

It is worth noting that, unlike the utility function $\util(\bx;\hat{\btheta})$ which depends on a changing model $\hat{\btheta}$, the explicit goal function $\goal(\btheta)$ in \goral{} serves a more comparable target across AL iterations. As a result it might be easier to measure progress as well as perform analyses in \goral{}, as we will demonstrate later. However, one should also note that $\util_{\text{goal}}$ can be much more expensive to compute due to its dependency on model \emph{retraining} (potentially $K$ times per sample $\bx$ due to $y$ being unknown), and to carry out such evaluations over a large pool of unlabelled samples only further adds to the problem. Fortunately, we show in Sec.~\ref{Sec:if} that with the help of influence functions recently leveraged by~\citep{koh17ubpif} for interpreting black-box model predictions, this high cost of model retraining could be drastically reduced to make \goral{} much more practical for a wide class of models.

\subsection{Choices of \texorpdfstring{$\ell_y$}{the operator} and \texorpdfstring{$\goal(\btheta)$}{the goal function} in \goral{}}
\label{Sec:ell_n_goal}
The different choices of $\ell_y$ and $\goal(\btheta)$ have lead to various AL strategies being proposed.

For $\ell_y$, we consider
\begin{enumerate*}[label={\alph*)},font={\bfseries}]
  \item \textbf{Expectation ($\expect_y$):} This is probably the most obvious choice, and one can take $P(y)$ directly from $P_{\hat{\btheta}}(y|\bx)$~\citep{roy01toal,schein07al4lr}, indirectly from a separately estimated ``oracle'' model~\citep{evans15mri}, or simply set it to a uniform distribution;
  \item \textbf{Min-/Max-imization ($\min_y$/$\max_y$):} This is being most pessimistic~\citep{hoi08semi} / optimistic~\citep{guo07oal} as it only looks at the extreme goal that could be achieved amongst all possible labels of $\bx$;
  \item \textbf{Oracle (``set $y$ to ground-truth''):} This is the ideal (albeit unrealistic) case which should provide a performance ``upper bound'' for the other choices to compare against.
\end{enumerate*}
Another possibility could be to borrow from the various \emph{acquisition functions} developed for Bayesian Optimization~\citep{mockus12bayesopt} although we keep to the above ones for the scope of this paper.

For $\goal(\btheta)$, we consider
\begin{enumerate*}[label={\alph*)},font={\bfseries}]
  \item \textbf{Negative dev-set loss ($\goal_{\text{dev}}$):} The negative cross-entropy loss on a held-out development set, i.e. $\goal_{\text{dev}}(\btheta;\mathcal{L}_{\text{dev}})\triangleq\sum_{(\bx,y)\in\mathcal{L}_{\text{dev}}}\log P_{\btheta}(y|\bx)$, serves a good proxy to model accuracy. However it is prone to \emph{over-fitting} and incurs extra labelling costs;
  \item \textbf{Negative prediction entropy ($\goal_{\text{ent}}$):} This can be thought of as the negative model uncertainty on an unlabelled dataset $\mathcal{U}$, i.e. $\goal_{\text{ent}}(\btheta;\mathcal{U})\triangleq-\sum_{\bx\in\mathcal{U}}\entropy(P_{\btheta}(y|\bx))$, where $\entropy$ stands for entropy. The rationale behind this goal is that one should favour models that are more certain about its predictions on unseen data~\citep{roy01toal,grandvalet04ent,guo07batch};
  \item \textbf{Negative Fisher information ($\goal_{\text{fir}}$):} “\textbf{F}isher \textbf{i}nformation \textbf{r}atio”~\citep{zhang00pavud} captures unlabelled samples' impact on the asymptotic efficiency of parameter estimation. More recently, \citep{sourati17fial} further shows it serves an asymptotic upper bound of the expected variance of the log-likelihood ratio. Inspired by these findings, we propose $\goal_{\text{fir}}(\btheta;\mathcal{U})\triangleq-\trace(\fisherinfo_{u}(\btheta))$, where $\fisherinfo_{u}(\btheta)\triangleq\frac{1}{|\mathcal{U}|}\sum_{\bx\in\mathcal{U}}\fisherinfo(\btheta|\bx)$ represents the empirical conditional Fisher information matrix (see Sec.~\ref{Supp:goal_fir} for details).
\end{enumerate*}

\section{Approximating \goral{} with influence functions}
\label{Sec:if}
As explained above, to accurately evaluate the goal-oriented utility per Eq.~\eqref{Eq:util_goal} over a large pool of unlabelled samples appears prohibitively expensive in general. In this section we show how, for a wide class of models, this can be efficiently approximated using influence functions~\citep{cook82if,koh17ubpif}.

Given a training set $\mathcal{L}_{\text{train}}=\{\bz_i\}_{i=1}^n$, we now assume the model $\hat{\btheta}$ is obtained via empirical risk minimization, i.e. {\small$\hat{\btheta}\triangleq\argmin_{\btheta}\frac{1}{n}\sum_{i=1}^n R(\bz_i, \btheta)$}, where $R$ is the per-sample loss function (with any regularization terms folded in). We then define {\small$\hat{\btheta}_{\epsilon, \bz}\triangleq\argmin_{\btheta}\frac{1}{n}\sum_{i=1}^n R(\bz_i, \btheta) + \epsilon R(\bz, \btheta)$} to be the new model trained with an additional $\epsilon$-weighted training sample $\bz$. Following~\citep{koh17ubpif}, under certain regularity conditions (e.g. $R$ being twice-differentiable and strictly convex), influence functions provide a closed-form estimate for the difference in model parameters $\hat{\btheta}_{\epsilon, \bz}-\hat{\btheta}$ (when $\epsilon$ is small) via $\left.\frac{\partial\hat{\btheta}_{\epsilon,\bz}}{\partial\epsilon}\right|_{\epsilon=0}=-H^{-1}_{\hat{\btheta}}\nabla_{\btheta}R(\bz,\hat{\btheta})$, where {\small $H_{\hat{\btheta}}\triangleq\frac{1}{n}\sum_{i=1}^n\nabla^2_{\btheta}R(\bz_i,\hat{\btheta})$} denotes the Hessian.

Following the chain rule and assuming $\goal(\btheta)$ is differentiable, we can now measure the ``influence'' of introducing an infinitesimally $\epsilon$-weighted sample $\bz$ (on the goal) as
\begin{align}
\label{Eq:inf_goal}
\mathcal{I}(\bz;\hat{\btheta})
\triangleq\frac{\partial\goal(\hat{\btheta}_{\epsilon,\bz})}{\partial\epsilon}\bigg|_{\epsilon=0}
=\frac{\partial\goal(\btheta)}{\partial\btheta}\bigg|_{\btheta=\hat{\btheta}}\cdot\frac{\partial\hat{\btheta}_{\epsilon,\bz}}{\partial\epsilon}\bigg|_{\epsilon=0}
=-\nabla^\top_{\btheta}\goal(\hat{\btheta})H^{-1}_{\hat{\btheta}}\cdot\nabla_{\btheta}R(\bz,\hat{\btheta})\text{,}
\end{align}
which we then leverage to form the approximation (using 1st-order Taylor approximation of $\goal(\hat{\btheta}_{\epsilon,\bz})$):
\begin{align}
\label{Eq:approx_util}
\util_{\text{goal}}(\bx;\hat{\btheta})=\ell_y\Big[\goal(\hat{\btheta}_{\epsilon,\bz})\big|_{\epsilon=\frac{1}{n}}\Big]-\goal(\hat{\btheta})\approx\ell_y\Big[\goal(\hat{\btheta})+\frac{1}{n}\mathcal{I}(\bz;\hat{\btheta})\Big]-\goal(\hat{\btheta})=\frac{1}{n}\ell_y\big[\mathcal{I}(\bz;\hat{\btheta})\big]\text{,}
\end{align}
and henceforth define $\tilde{\util}_{\text{goal}}(\bx;\hat{\btheta})\triangleq\frac{1}{n}\ell_y\big[\mathcal{I}(\bz;\hat{\btheta})\big]$ to be the approximate goal-oriented utility.\footnote{The last equation in Eq.~\eqref{Eq:approx_util} holds for all the $\ell_y$ operators considered in this paper (see Sec.~\ref{Sec:ell_n_goal}).}

Note that when $\ell_y$ is a linear operator (e.g. Expectation or Oracle), it can switch order with $\nabla_\epsilon$, i.e. $\nabla_{\epsilon}\ell_y[\goal(\hat{\btheta}_{\epsilon,\bz})]=\ell_y[\nabla_{\epsilon}\goal(\hat{\btheta}_{\epsilon,\bz})]$, and as a result, $\ell_y[\goal(\hat{\btheta}_{\epsilon,\bz})]-\goal(\hat{\btheta})=\epsilon\cdot\ell_y[\mathcal{I}(\bz;\hat{\btheta})]+o(\epsilon)$, making $\tilde{\util}_{\text{goal}}(\bx;\hat{\btheta})$ itself a direct 1st-order Taylor approximation to $\util_{\text{goal}}(\bx;\hat{\btheta})$. Otherwise (e.g. for $\max_y$), Eq.~\eqref{Eq:approx_util} may offer a looser approximation, although we find it still works well in practice.

Also note it is obvious from the r.h.s. of Eq.~\eqref{Eq:inf_goal} that $\tilde{\util}_{\text{goal}}(\bx;\hat{\btheta})$ can be further broken down into two terms, namely $\tilde{\util}_{\text{goal}}(\bx;\hat{\btheta})=\ell_y\big[\bv^\top_{\hat{\btheta}}\cdot\nabla_{\btheta}R(\bz,\hat{\btheta})\big]$, where $\bv_{\hat{\btheta}}\triangleq-\frac{1}{n}H^{-1}_{\hat{\btheta}}\nabla_{\btheta}\goal(\hat{\btheta})$ is independent of $\bz$ and hence, once computed, can be reused across all samples. Therefore, computation-wise, even though to evaluate $\util_{\text{goal}}(\bx;\hat{\btheta})$ over $\mathcal{U}_{\text{pool}}$ requires $K\times|\mathcal{U}_{\text{pool}}|$ iterations of model retraining, for $\tilde{\util}_{\text{goal}}(\bx;\hat{\btheta})$ it now only requires the same number of gradient computations, i.e. $\nabla_{\btheta}R(\bz,\hat{\btheta})$.

\subsection{Extra caution required for the Expectation operator}
\label{Sec:cpy}
Nice and intuitive as it sounds, below we present a perhaps surprising result for the Expectation operator $\expect_y$, which is that the popular choice of taking the expectation under the current model prediction for resolving the unknown label $y$ may actually render the resulting utility vacuous.
\begin{remark}
\label{Rmk:exp_const}
With (regularized) maximum-likelihood estimate (or cross-entropy loss), i.e. whenever $R(\bz,\btheta)=\Omega(\btheta)-\log P_{\btheta}(y|\bx)$, the approximate expected utility $\tilde{\util}_{\text{exp}}(\bx;\hat{\btheta})\triangleq\frac{1}{n}\expect_{y}\big[\mathcal{I}(\bz;\hat{\btheta})\big]$ under the current model prediction (i.e. $y\sim P_{\hat{\btheta}}(y|\bx)$) becomes a constant regardless of the sample $\bx$.
\end{remark}
To see why, just note that
\begin{align}
\tilde{\util}_{\text{exp}}(\bx;\hat{\btheta})&=-\frac{1}{n}\nabla^\top_{\btheta}\goal(\hat{\btheta})H^{-1}_{\hat{\btheta}}\cdot\expect_y[\nabla_{\btheta}R(\bz,\hat{\btheta})]\nonumber\\
&=-\frac{1}{n}\nabla^\top_{\btheta}\goal(\hat{\btheta})H^{-1}_{\hat{\btheta}}\cdot\left(\expect_y[\nabla_{\btheta}\Omega(\hat{\btheta})]-\expect_y[\nabla_{\btheta}\log P_{\hat{\btheta}}(y|\bx)]\right)\nonumber\\
&=-\frac{1}{n}\nabla^\top_{\btheta}\goal(\hat{\btheta})H^{-1}_{\hat{\btheta}}\cdot\nabla_{\btheta}\Omega(\hat{\btheta})\equiv\text{const.,}\nonumber
\end{align}
where the last step naturally follows from the well known result that the \emph{score function} has zero mean, i.e. $\expect_{y}[\nabla_{\btheta}\log P_{\btheta}(y|\bx)]=\bm{0}$.

It is also worth noting that the above remark holds regardless of the choice of the goal function $\goal(\btheta)$. And therefore for model classes fitting the above assumptions, e.g. logistic regression, this seems a rather poor choice \textemdash{} the actual utility is both expensive and susceptible to noises, while its 1st-order Taylor-approximate utility turns out unusable still.

\subsection{\goral{} in batch-mode}
In batch-mode AL, instead of just selecting the top one sample, multiple samples are to be selected, labelled, and then added to the training set in one go at every iteration. Doing so facilitates \emph{less greedy} AL strategies as it allows one to evaluate and process a batch as a whole (e.g. to take diversity into consideration) rather than sticking to successive locally-optimal individual selections.

However, this is not exempt from the ``no free lunch'' principle. Denote a batch by $X\triangleq\{\bx\}$. Due to the combinatorial nature of subset selection, optimizing a holistic batch utility $\util(X;\hat{\btheta})$ typically results in a much higher computational cost (e.g. one which scales exponentially with the batch size). As a result, many batch-mode AL strategies in practice \emph{choose to} simply compose its batch utility from the sum of individual utilities, i.e. $\util(X;\hat{\btheta})=\sum_{\bx\in X}\util(\bx;\hat{\btheta})$. However, doing so essentially reduces it back to the greedy setting and seems like a heuristic at best. For batch-mode \goral{} though, we can actually enjoy the best of both worlds \textemdash{} the definition of the goal-oriented utility~\eqref{Eq:util_goal} naturally lends itself to a holistic batch-mode version, yet, as we show below, it still benefits from cheap computation through principled approximations.

Thanks to the explicit goal function $\goal(\btheta)$, we naturally extend the definition of the utility (Eq.~\eqref{Eq:util_goal}) to the batch-mode setting by following the same principle, i.e. $\util_{\text{goal}}(X;\hat{\btheta})\triangleq\ell_{Y}\big[\goal(\hat{\btheta}_{Z})\big]-\goal(\hat{\btheta})$, where $Z$ denotes the batch augmented with hypothetical labels $Y$, $\hat{\btheta}_{Z}$ the model trained with this additional (labelled) batch, and $\ell_Y$ the operator that resolves the unknown $Y$ (similar to Sec.~\ref{Sec:ell_n_goal}).

Similarly, we use $\hat{\btheta}_{\epsilon, Z}\triangleq\argmin_{\btheta}\frac{1}{n}\sum_{i=1}^n R(\bz_i, \btheta) + \frac{\epsilon}{b}\sum_{\bz\in Z}R(\bz, \btheta)$ ($b$ being the batch size $|Z|$) to study the ``influence'' of introducing a batch of samples $Z$, and as it turns out,
\begin{align}
\mathcal{I}(Z;\hat{\btheta})\triangleq\frac{\partial\goal(\hat{\btheta}_{\epsilon,Z})}{\partial\epsilon}\bigg|_{\epsilon=0}=-\nabla^\top_{\btheta}\goal(\hat{\btheta})H^{-1}_{\hat{\btheta}}\cdot\frac{1}{b}\sum_{\bz\in Z}\nabla_{\btheta}R(\bz, \hat{\btheta})=\frac{1}{b}\sum_{\bz\in Z}\mathcal{I}(\bz;\hat{\btheta})\text{,}\nonumber
\end{align}
which means the \emph{collective} influence $\mathcal{I}(Z;\hat{\btheta})$ is simply the \emph{average} of the individual $\mathcal{I}(\bz;\hat{\btheta})$s.

Then applying the same approximation idea as above (Eq.~\eqref{Eq:approx_util}) and assuming $b \ll n$, we have
\begin{align}
\util_{\text{goal}}(X;\hat{\btheta})=\ell_{Y}\Big[\goal(\hat{\btheta}_{\epsilon,Z})\big|_{\epsilon=\frac{b}{n}}\Big]-\goal(\hat{\btheta})\approx\ell_{Y}\Big[\goal(\hat{\btheta})+\frac{b}{n}\mathcal{I}(Z;\hat{\btheta})\Big]-\goal(\hat{\btheta})=\frac{b}{n}\ell_{Y}\big[\mathcal{I}(Z;\hat{\btheta})\big]\text{,}\nonumber
\end{align}
and thus denote $\tilde{\util}_{\text{goal}}(X;\hat{\btheta})\triangleq\frac{b}{n}\ell_{Y}\big[\mathcal{I}(Z;\hat{\btheta})\big]$ to be the approximate batch utility.

\begin{remark}
\label{Rmk:batch_sum}
The approximate batch utility is the same as the sum of the approximate individual utilities, i.e. $\tilde{\util}_{\text{goal}}(X;\hat{\btheta})=\sum_{\bx\in X}\tilde{\util}_{\text{goal}}(\bx;\hat{\btheta})$.
\end{remark}
This is straightforward as {\small$\frac{b}{n}\ell_Y\big[\mathcal{I}(Z;\hat{\btheta})\big]=\ell_{Y}\big[\frac{1}{n}\sum_{\bz\in Z}\mathcal{I}(\bz;\hat{\btheta})\big]=\sum_{\bx\in X}\frac{1}{n}\ell_y\big[\mathcal{I}(\bz;\hat{\btheta})\big]$}.\footnote{The 2nd equation holds for all the $\ell_y$ operators considered in Sec.~\ref{Sec:ell_n_goal} in that their extension $\ell_Y$s are all \emph{decomposable}, i.e. $\ell_Y\big[\sum_{y\in Y}f(y)\big]=\sum_{y\in Y}\ell_y[f(y)]$.} Remark~\ref{Rmk:batch_sum} implies that using greedy selection for batch-mode \goral{} is actually well-justified.

Computation-wise, accurately selecting the best batch (of size $b$) from the pool requires $\binom{|\mathcal{U}_{\text{pool}}|}{b}$ times of $\util_{\text{goal}}(X;\hat{\btheta})$ evaluations, each of which in turn requires $K^b$ times of model retraining, both scaling \emph{exponentially} with the batch size $b$. Under the approximation, this cost gets drastically reduced to $K\times|\mathcal{U}_{\text{pool}}|$ times of gradient computations, plus a one-time top-$b$ item selection from the pool.

\section{Discussion}
In Sec.~\ref{Supp:exp} we present empirical studies that convince the effectiveness of the proposed approximation (\ref{supp:approx_quality}), showcase \goral{}’s robustness against an adversarial setting (\ref{supp:synth2}), as well as highlight worrying problems with all the three goal functions considered in Sec.~\ref{Sec:ell_n_goal} with honest benchmark results on two representative, non-cherrypicked datasets (\ref{supp:benchmark}). In particular, we demonstrate the close relationship between $\goal_{\text{ent}}$ and $\goal_{\text{fir}}$ (also analytically in Sec.~\ref{Sec:logreg}), and that both are a poor goal for AL as achieving the goal is actually at odds with achieving good data efficiency. In this regard, $\goal_{\text{dev}}$ performs much better but it needs to address its own issues of potential over-fitting and extra labelling costs.

Further to that, the analytical insights presented in Sec.~\ref{Sec:cpy} challenge the status quo and prompt us to seek better alternatives when choosing to take expectation over the labels for utility estimation.

\bibliographystyle{unsrtnat}
\bibliography{references}

\newpage

\section{Discussion (continued)}

Since its introduction to machine learning, influence functions have also been successfully applied to the setting of ``optimal subsampling''~\citep{ting18sampling}, which bears some resemblance to active learning in that both are trying to select a subset from the data. However, the differences between these two settings are also stark and clear. In particular, for active learning, both its unique dependency on the unknown labels and the discrepancy between the proxy goal and the training objective call for more careful treatment, as have been demonstrated in this paper.

When we were discussing the computational cost of \goral{} and its approximation in Sec.~\ref{Sec:if}, our primary focus was on those terms that either scale with the size of the pool $|\mathcal{U}_{\text{pool}}|$ or the batch $b$, since they are the more dominant ones (especially in batch-mode). Notably, the one-time cost of computing $\bv_{\hat{\btheta}}$ is also not to be neglected as it involves a Hessian ($O(|\mathcal{L}_{\text{train}}|d^2)$) and an inverse-Hessian-vector product. However, we note that, in the AL setting, it is expected that $|\mathcal{U}_{\text{pool}}|\gg|\mathcal{L}_{\text{train}}|$, and therefore the reduction in computational cost is still significant.

In regards to the issue of a vacuous utility resulting from directly using the current model prediction in the expectation (Sec.~\ref{Sec:cpy}), one possible remedy might be to \emph{soften} the prediction distribution~\citep{hinton15distill} by annealing it with a temperature $T\in\mathbb{R}^{+}$, i.e. setting $P(y)\propto\exp(\nicefrac{\log P_{\hat{\btheta}}(y|\bx)}{T})$. Note that this reduces it to a uniform distribution when $T\to\infty$, the original distribution when $T=1$, and a singleton distribution at $\argmax_y P_{\hat{\btheta}}(y|\bx)$ when $T\to 0$. Hence one can start with a relatively high temperature at the early stage of an AL process when the model is less well trained, and then progressively tune the temperature down as the model has been trained with more data and gets more accurate over time. We leave this to future work.

\section{Experiments}
\label{Supp:exp}
We now carry out empirical studies to validate the efficacy, as well as showcase some problems, of several \goral{} strategies. For this we use three datasets, i.e. \synth~\citep{huang10quire,yang18benchmark}, which is a binary classification dataset crafted to highlight issues with those AL strategies that focus on exploiting ``informative'' samples only (e.g. uncertainty sampling), \href{http://www.cs.cornell.edu/people/pabo/movie-review-data/rt-polaritydata.README.1.0.txt}{\rtpolar}~\citep{pang05rtpolar}, a binary sentence classification dataset, and \href{https://www.csie.ntu.edu.tw/~cjlin/libsvmtools/datasets/multiclass.html#letter}{\letter}~\citep{frey91letter}, a multi-class image classification dataset. For \rtpolar, we encode every sentence by taking its ``\texttt{[CLS]}'' embedding from \href{https://github.com/google-research/bert/blob/master/extract_features.py}{BERT}~\citep{devlin18bert}.

\begin{minipage}{\textwidth}
  \centering
  \begin{minipage}[b]{0.59\textwidth}
    \centering
    \captionsetup{type=table}
    \begin{tabular}{l|r|r|r|r|r}
      Dataset & $K$ & $d$ & $|\mathcal{U}_{\text{pool}}|$ & $|\mathcal{L}_{\text{init}}|$ & $|\mathcal{L}_{\text{test}}|$ \\
      \hline
      \synth & $2$ & $2$ & $530$ & $10$ & $60$ \\
      \rtpolar & $2$ & $768$ & $9,586$ & $10$ & $1,066$ \\
      \letter & $26$ & $16$ & $14,948$ & $52$ & $5,000$ \\
      \hline
    \end{tabular}
    \caption{Datasets}
    \label{Tab:datasets}
  \end{minipage}
  \hfill
  \begin{minipage}[b]{0.35\textwidth}
    \centering
    \captionsetup{type=figure}
    \begin{subfigure}{.49\textwidth}
      \includegraphics[width=\linewidth]{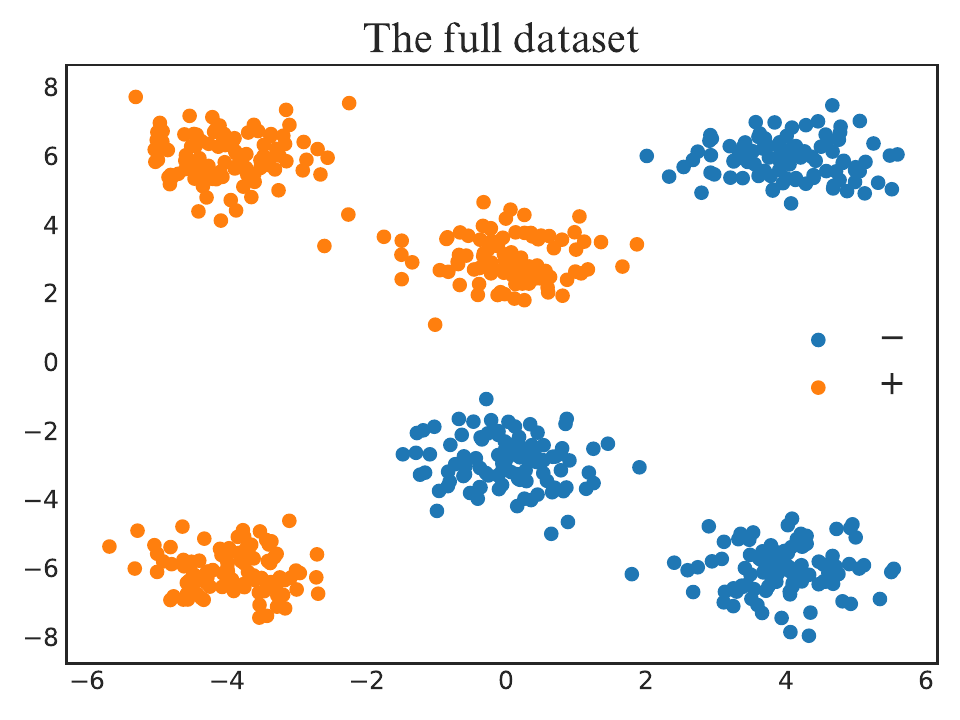}
    \end{subfigure}
    \hfill
    \begin{subfigure}{.49\textwidth}
      \includegraphics[width=\linewidth]{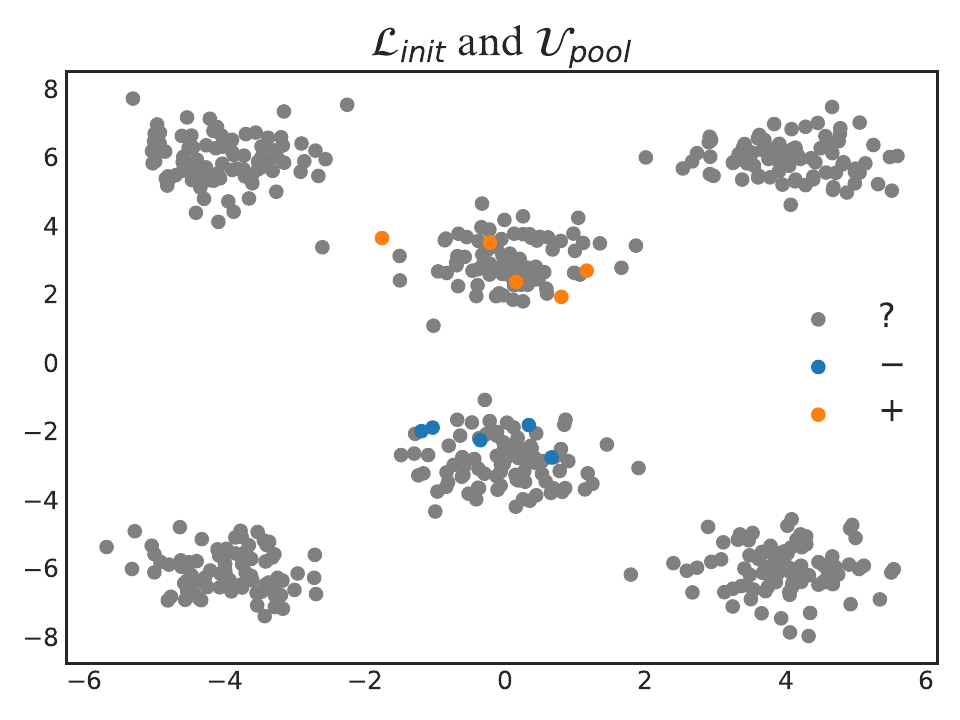}
    \end{subfigure}
    \caption{The \synth{} dataset}
    \label{fig:dataset_synth2}
  \end{minipage}
\end{minipage}

We focus on Multinomial Logistic Regression (MLR) for all the experiments. There has been a lot of research specifically concentrated on AL for logistic regression~\citep{schein07al4lr,yang18benchmark}. Furthermore, with the recent advent of powerful pre-trained models~\citep{devlin18bert}, it is becoming ever more promising that, by simply stacking an additional final layer (typically MLR for classification) on top of those pre-trained networks and \emph{fine-tuning} that layer's parameters to the given task one can readily obtain well performing models with little work. We include intercepts in the model and select the hyperparameter $\lambda$ with cross validation.\footnote{We provide a proper exposition along with all the derivations and details in Sec.~\ref{Sec:logreg}.}

\subsection{Approximation quality}
\label{supp:approx_quality}
\begin{figure}[!htb]
  \centering
  \begin{subfigure}{.245\textwidth}
    \centering
    \includegraphics[width=\linewidth]{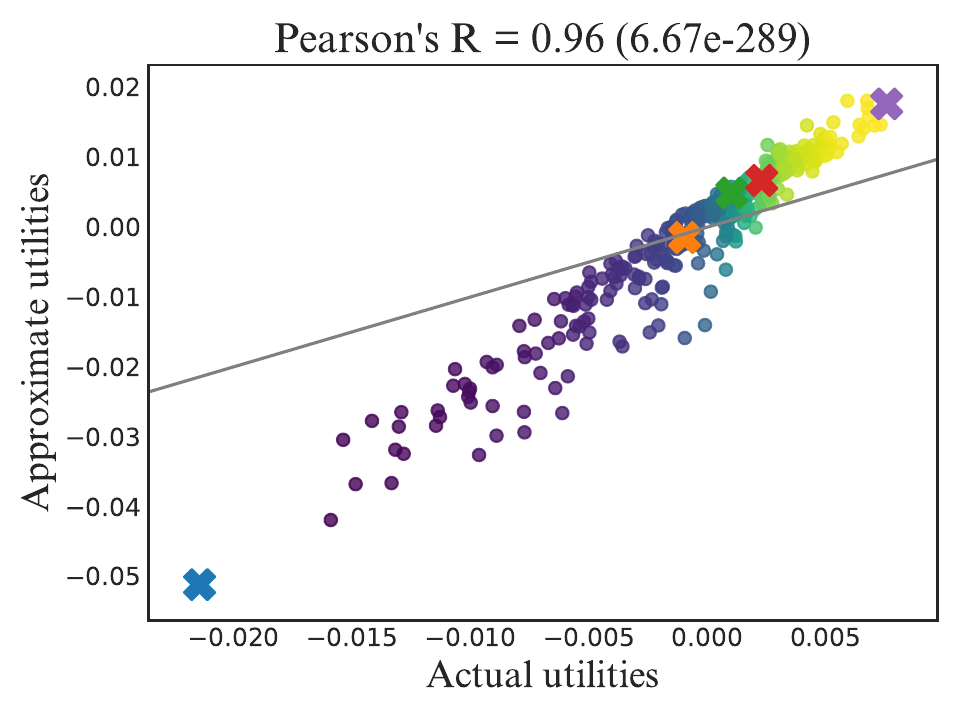}
    \caption{Oracle}
  \end{subfigure}
  \hfill
  \begin{subfigure}{.245\textwidth}
    \centering
    \includegraphics[width=\linewidth]{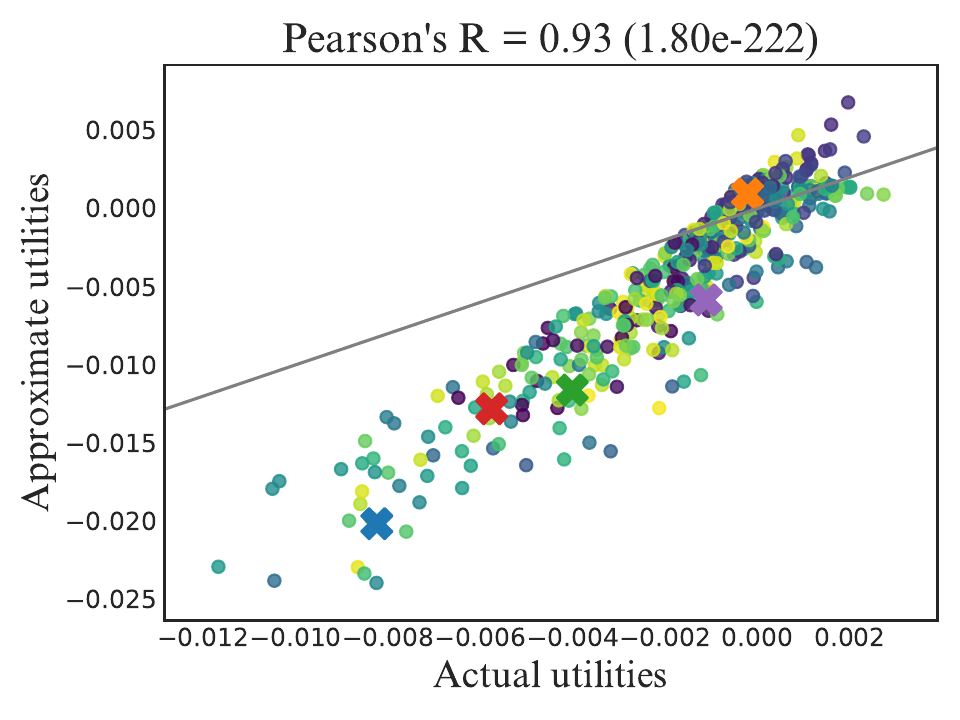}
    \caption{Average}
  \end{subfigure}
  \hfill
  \begin{subfigure}{.245\textwidth}
    \centering
    \includegraphics[width=\linewidth]{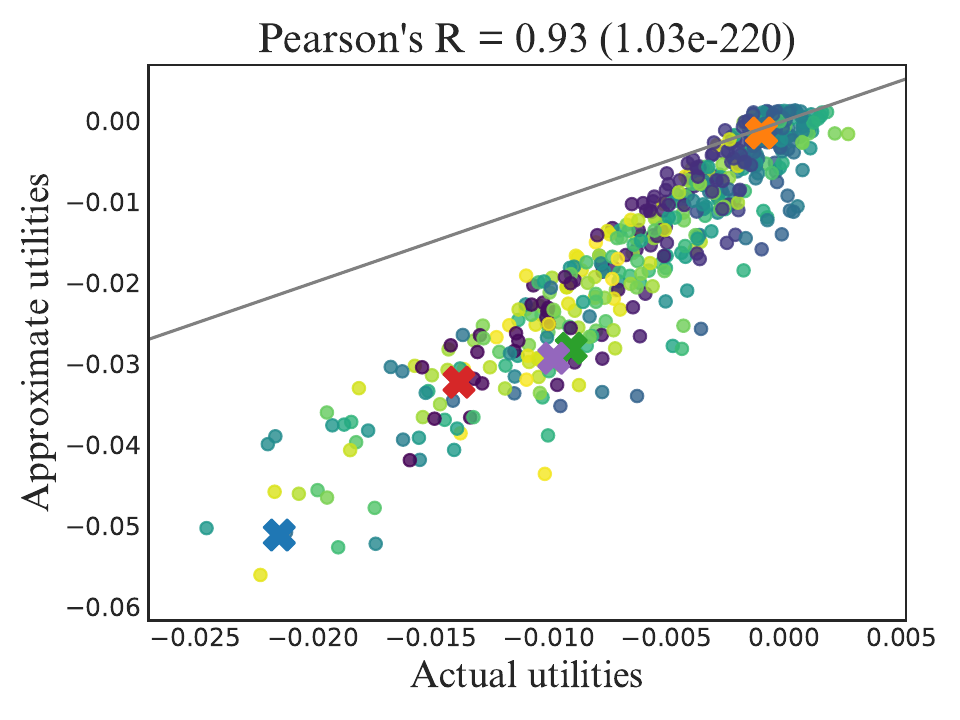}
    \caption{Minimum}
  \end{subfigure}
  \hfill
  \begin{subfigure}{.245\textwidth}
    \centering
    \includegraphics[width=\linewidth]{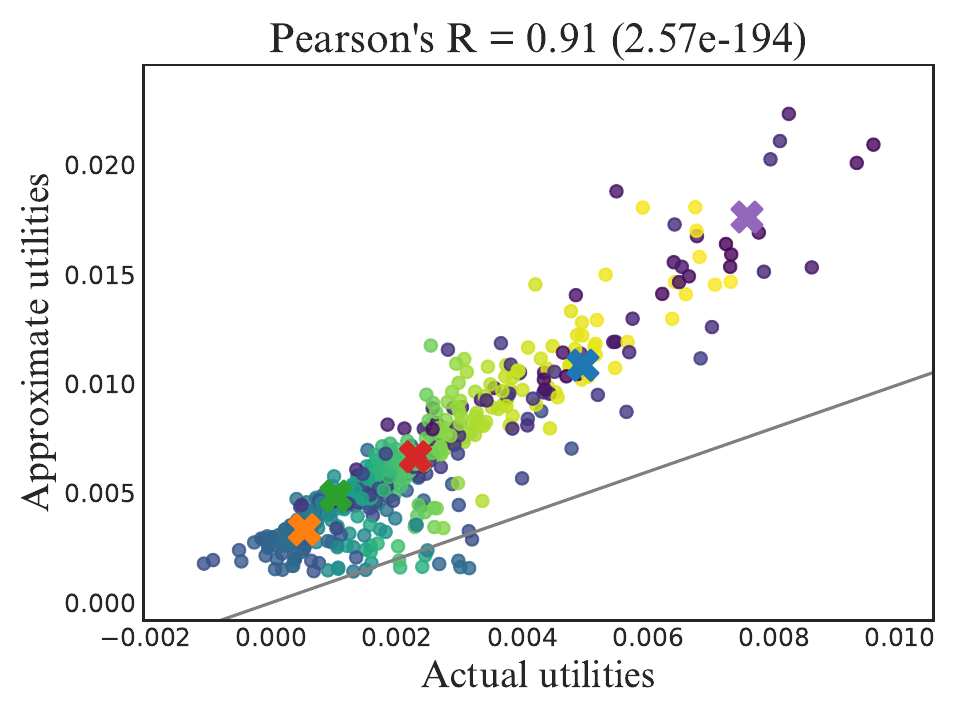}
    \caption{Maximum}
  \end{subfigure}
  \caption{Approximation quality of $\tilde{\util}_{\text{goal}}(\bx;\hat{\btheta})$ in the serial \goral{} setting under various $\ell_y$s.\newline
  Each dot represents one sample $\bx$ in the pool and is coloured consistently across plots (indexing into a color-map using its actual Oracle utility). The grey line represents the line $y=x$ for reference. And the crosses mark samples we single out for closer inspection.}
  \label{fig:approx_check_b1}
\end{figure}

We first examine how well the approximate utility proposed in Sec.~\ref{Sec:if} actually reflects the true utility. Here we present results on the goal function $\goal_{\text{ent}}$ only, since other goals all result in similar observations. We use the \rtpolar{} dataset and first train a model $\hat{\btheta}$ from $50$ random samples. We then compute both the actual utilities $\util_{\text{goal}}(\bx;\hat{\btheta})$ (by performing actual model retraining) and the approximate utilities $\tilde{\util}_{\text{goal}}(\bx;\hat{\btheta})$ (per Eq.~\eqref{Eq:approx_util}) over a pool of another $500$ random samples. From the scatter plots in Fig.~\ref{fig:approx_check_b1}, we see that overall the approximation works fairly well across all the various $\ell_y$s considered in this paper. Another observation is that the various $\ell_y$s do result in very different rankings among the samples, as is exemplified by the $5$ samples marked with crosses.

\begin{figure}[!htb]
  \centering
  \begin{subfigure}{.245\textwidth}
    \centering
    \includegraphics[width=\linewidth]{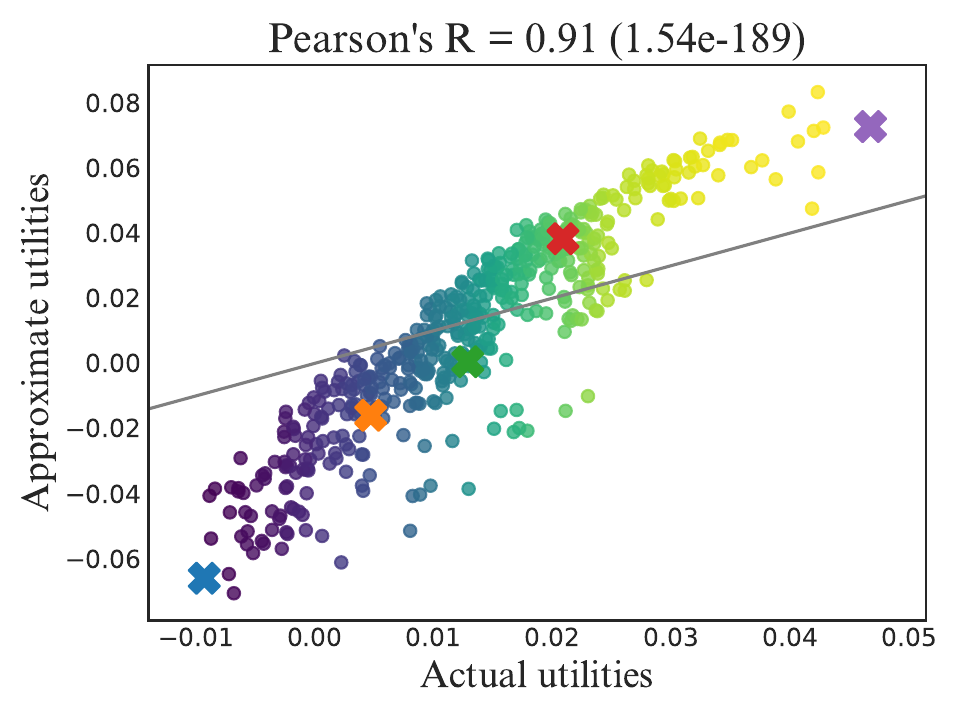}
    \caption{Oracle}
  \end{subfigure}
  \hfill
  \begin{subfigure}{.245\textwidth}
    \centering
    \includegraphics[width=\linewidth]{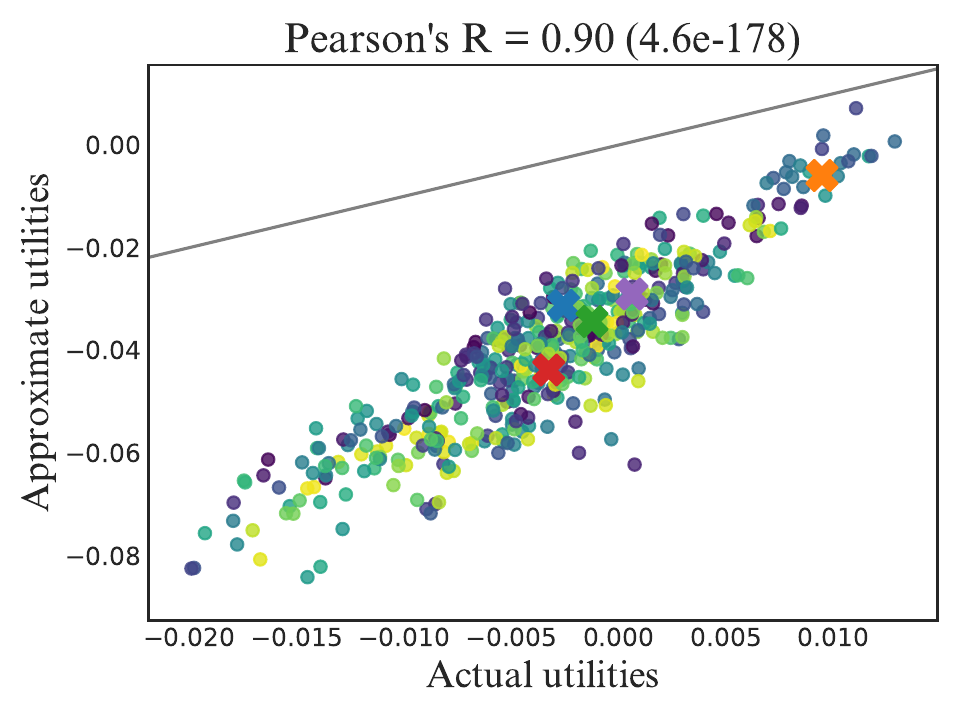}
    \caption{Average}
  \end{subfigure}
  \hfill
  \begin{subfigure}{.245\textwidth}
    \centering
    \includegraphics[width=\linewidth]{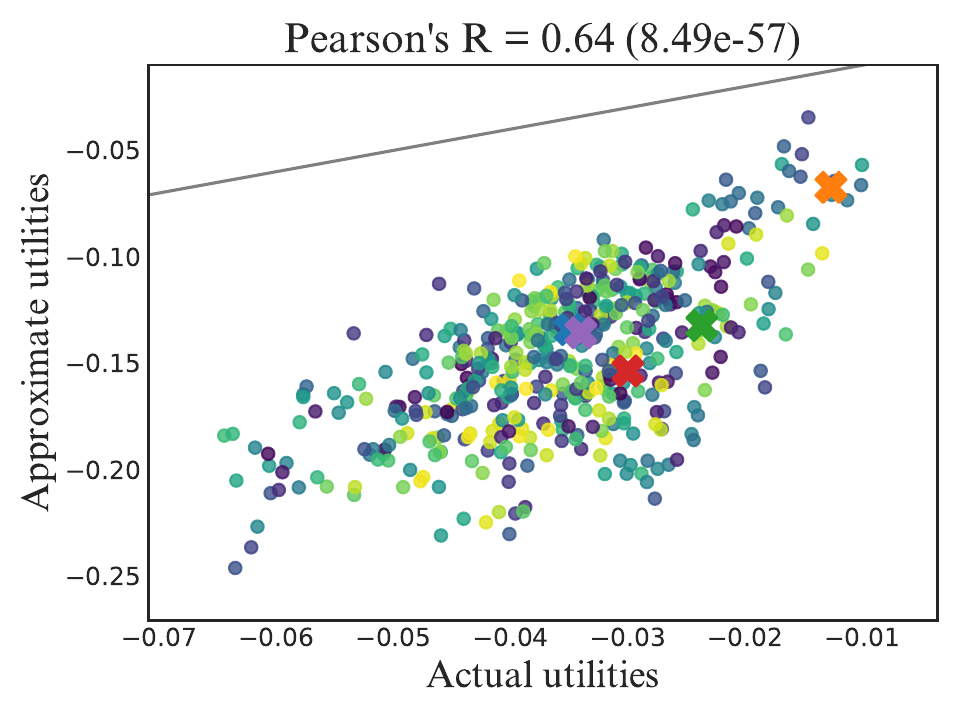}
    \caption{Minimum}
  \end{subfigure}
  \hfill
  \begin{subfigure}{.245\textwidth}
    \centering
    \includegraphics[width=\linewidth]{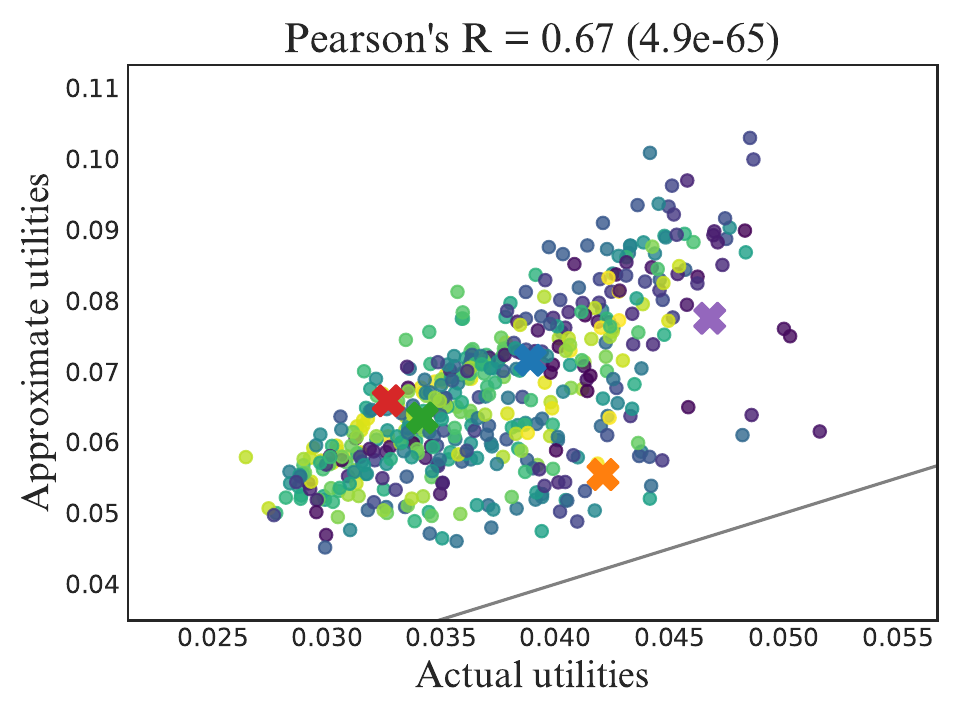}
    \caption{Maximum}
  \end{subfigure}
  \setlength{\belowcaptionskip}{-10pt}
  \caption{Approximation quality of $\tilde{\util}_{\text{goal}}(X;\hat{\btheta})$ in batch-mode \goral{} ($b=10$) under various $\ell_y$s. All elements are similar to those in Fig.~\ref{fig:approx_check_b1} except that each dot (or cross) now represents a batch $X$.}
  \label{fig:approx_check_b10}
\end{figure}

In Fig.~\ref{fig:approx_check_b10} we examine the same approximation quality for the batch-mode setting (with batch size $b=10$). For practical reasons, we don't examine all the possible $\binom{500}{10}\approx2.5\times10^{20}$ batches, but instead just pick $491$ batches from the same pool using a sliding window. Compared to the serial setting, we observe a slight degradation of the approximation quality when $\ell_y$ is linear (i.e. Expectation or Oracle), and a larger degradation when $\ell_y$ is $\max_y$ or $\min_y$, echoing our analysis in Sec.~\ref{Sec:if}. Nonetheless, in all cases the approximate utilities still exhibit a strong correlation with the actual ones (even when $\nicefrac{b}{n}$ is as high as $\nicefrac{10}{50}=0.2$), which is an inspiring result.

Concurrently, \citep{koh19oaifmge} studies the accuracy of influence function approximations for measuring group effects and makes similar observations that the approximation quality is generally very high although it can consistently under-estimate (over-estimate in our case).

\begin{figure}[!htb]
  \centering
  \begin{subfigure}{.66\textwidth}
    \centering
    \includegraphics[width=.325\linewidth]{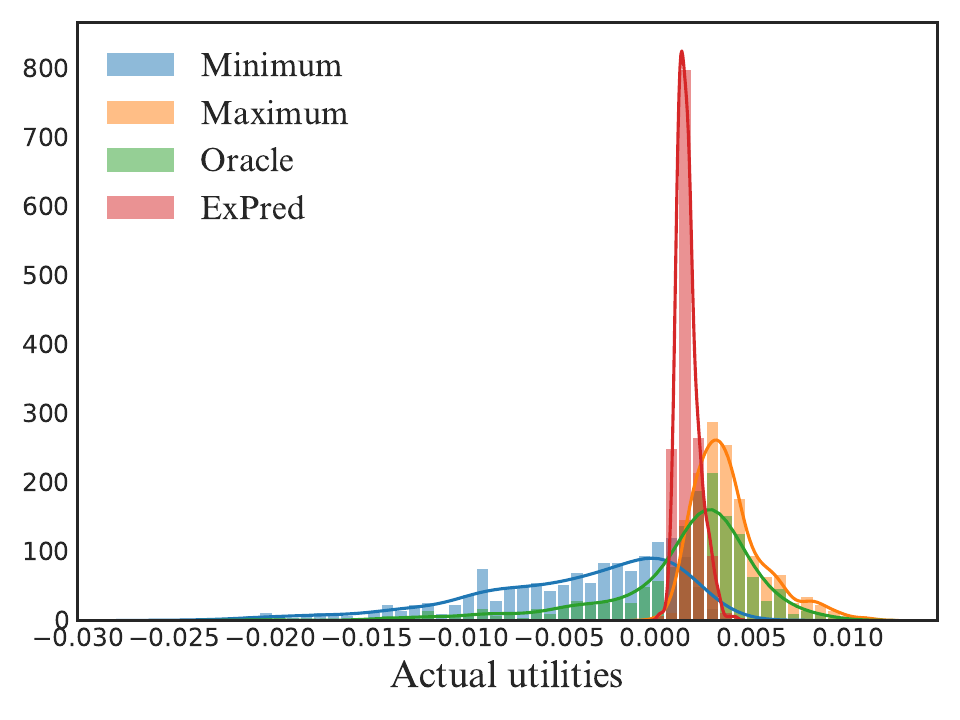}
    \hfill
    \includegraphics[width=.325\linewidth]{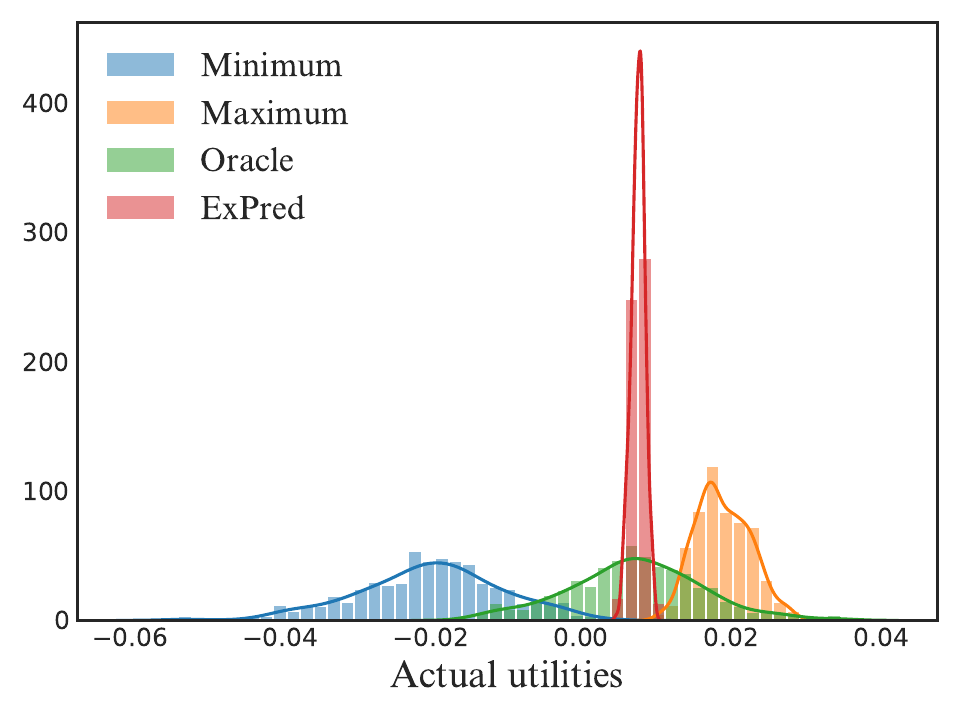}
    \hfill
    \includegraphics[width=.325\linewidth]{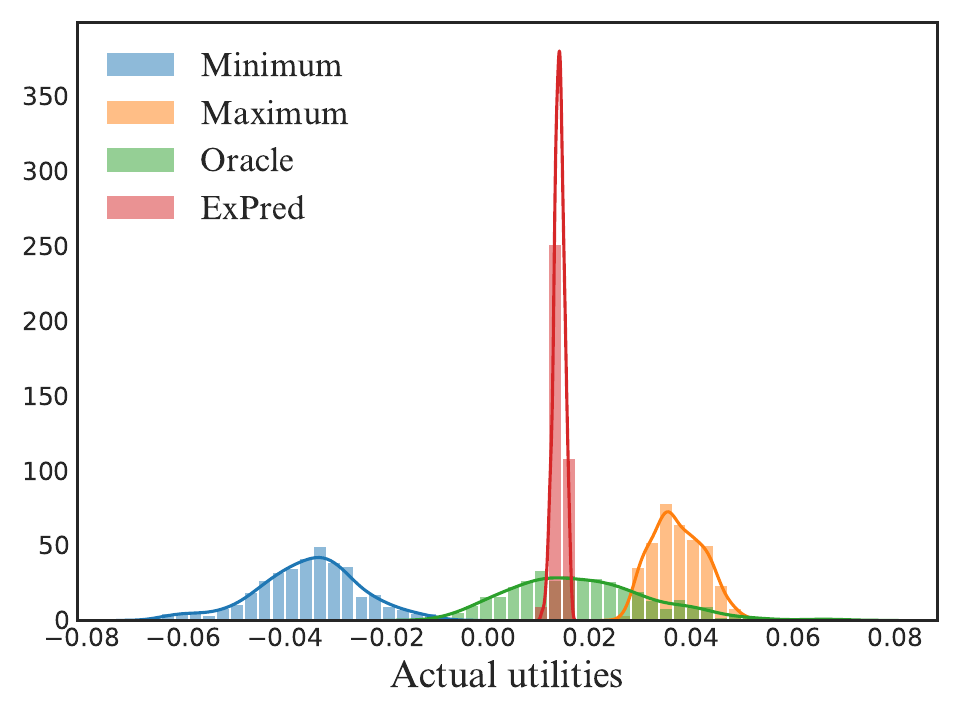}
    \caption{$\goal_{\text{ent}}$ with batch size $1$ (left), $5$ (middle) and $10$ (right).}
  \end{subfigure}
  \hfill
  \begin{subfigure}{.32\textwidth}
    \centering
    \includegraphics[width=.675\linewidth]{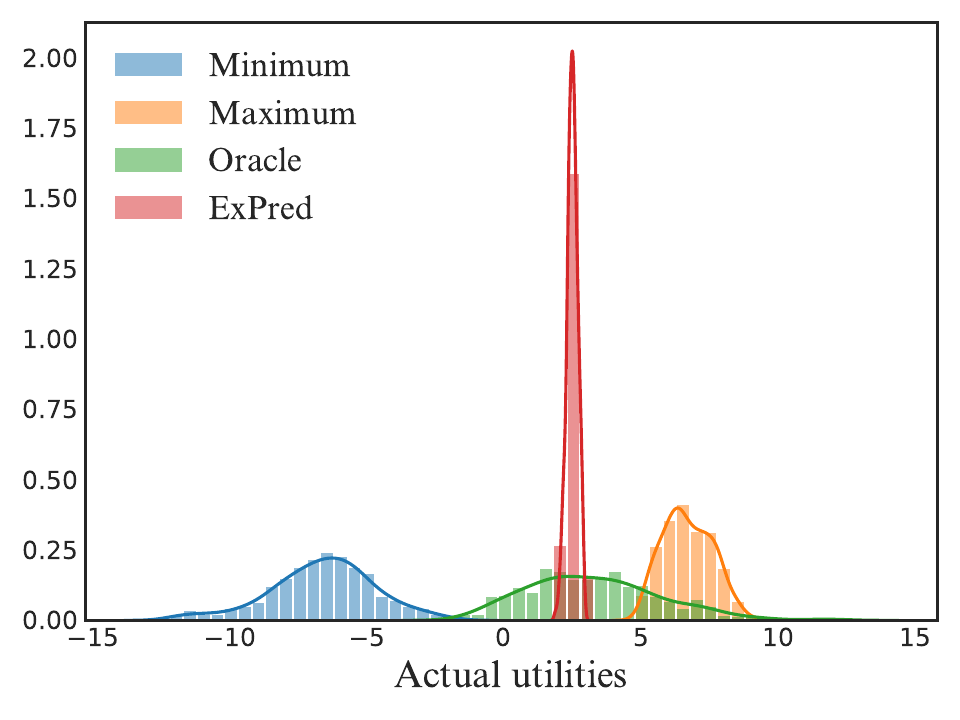}
    \caption{$\goal_{\text{fir}}$ with batch size $10$.}
  \end{subfigure}
  \caption{Utility distributions under various goals and batch sizes.}
  \label{fig:cpy_util}
\end{figure}

For the special case of $\expect_{y\sim P_{\hat{\btheta}}(y|\bx)}$, i.e. expectation under the current model prediction, we have known from Sec.~\ref{Sec:cpy} that the approximate utilities will be constant. In Fig.~\ref{fig:cpy_util} we show the histogram, along with the kernel density estimate, of the actual utilities under this operator (legend ``ExPred''), and contrast it with utilities from some other operators, across various batch sizes. From this we see that utilities under $\expect_{y\sim P_{\hat{\btheta}}(y|\bx)}$ are indeed highly concentrated within a fairly small region, making it highly susceptible to noise (e.g. due to training or numerical instabilities) and therefore less meaningful as a reliable criterion for sample selection in \goral{}.

In the following subsections, we benchmark several representative \goral{} strategies on a series of datasets along with baselines such as random sampling and uncertainty sampling, which are recently shown to be the more consistently effective AL strategies in~\citep{yang18benchmark}. And when we mention \goral{}, we now always refer to its practical version using influence-function approximations and use its batch-mode version with $b=10$.

\subsection{\synth: an adversarial setting}
\label{supp:synth2}
From Fig.~\ref{fig:dataset_synth2} we see that $\mathcal{L}_{\text{init}}$ is deliberately crafted to mislead the initial model, as we would like to inspect how \emph{robust} an AL strategy is by checking how quickly it can recover from that. In particular, the \synth{} dataset is composed of three groups of clusters, which we name as ``central'', ``distracting'', and ``definitive'' respectively. Within each group there are always two clusters, one for the positive label and one for the negative. The two central clusters are where $\mathcal{L}_{\text{init}}$ is drawn, and are poised to mislead the initial model into a nearly horizontal divide between the two; The two distracting clusters lie on the upper-left and lower-right corners, and are composed of samples with the largest distances to the ground-truth decision boundary; While the two definitive clusters lie on the upper-right and lower-left corners and, along with the two central clusters, define the optimal decision boundary.

For \goral{} we look at $\goal_\text{dev}$, where the dev-set $\mathcal{L}_{\text{dev}}$ is composed of a random $10$\% subset of $\mathcal{U}_{\text{pool}}$ (i.e. $53$ samples) associated with their ground-truth labels.

\begin{figure}[!htb]
  \centering
  \begin{subfigure}{\textwidth}
    \centering
    \includegraphics[width=.245\linewidth]{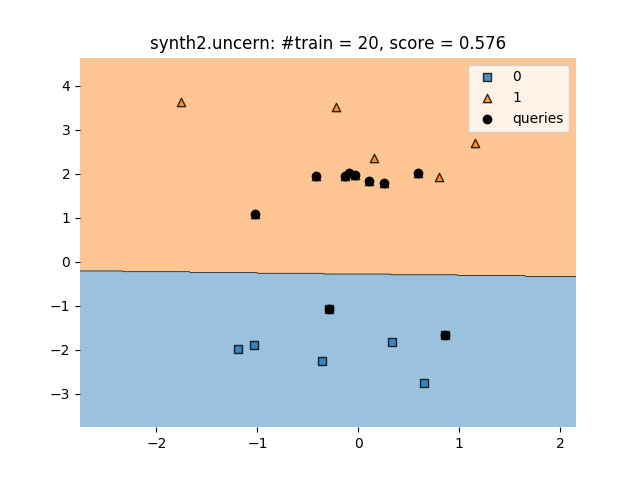}
    \hfill
    \includegraphics[width=.245\linewidth]{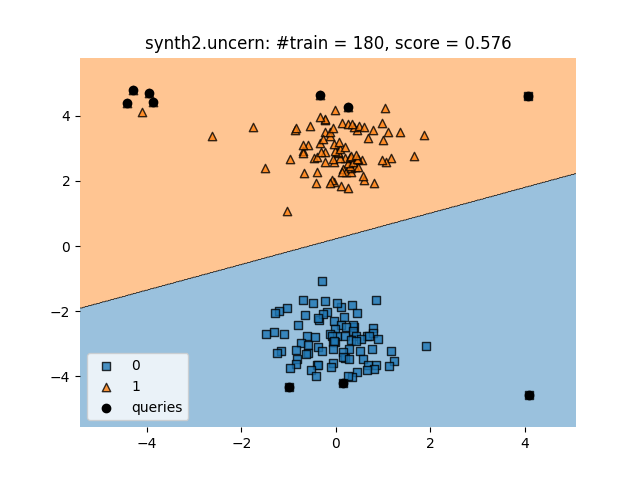}
    \hfill
    \includegraphics[width=.245\linewidth]{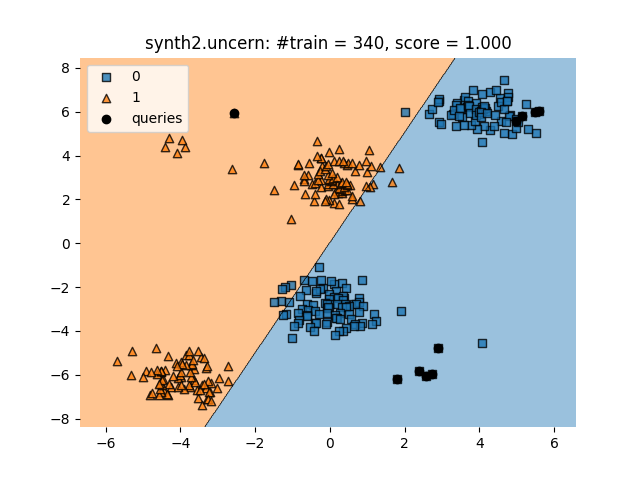}
    \hfill
    \includegraphics[width=.245\linewidth]{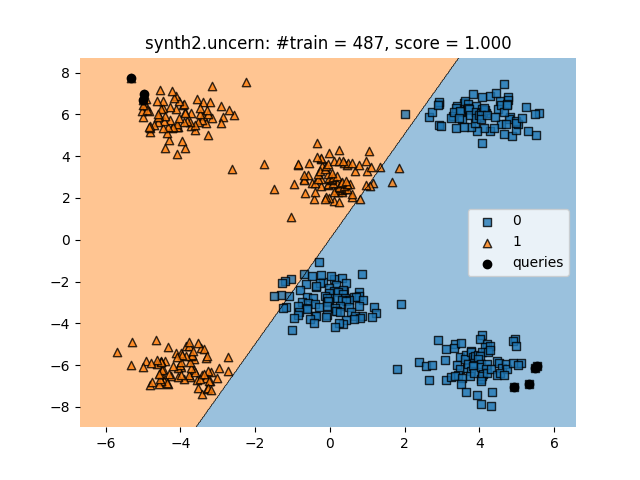}
    \caption{Uncertainty sampling}
  \end{subfigure}

  \begin{subfigure}{\textwidth}
    \centering
    \includegraphics[width=.245\linewidth]{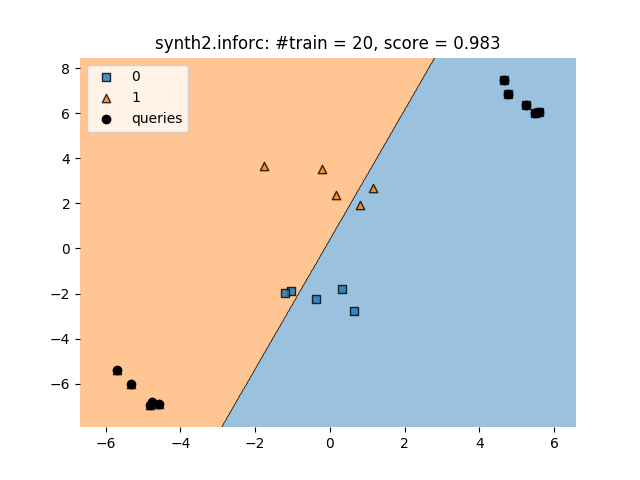}
    \hfill
    \includegraphics[width=.245\linewidth]{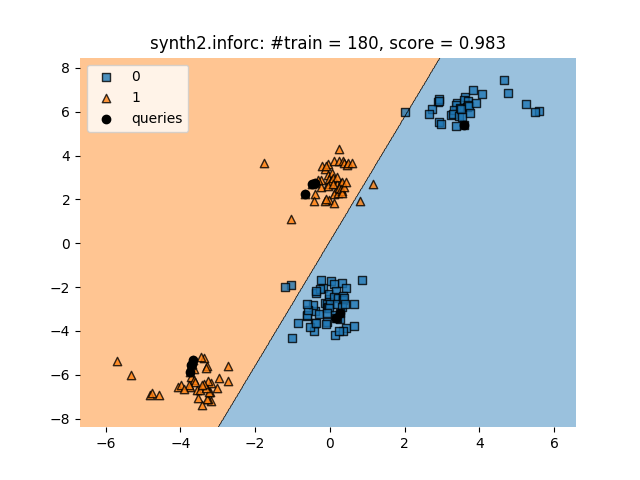}
    \hfill
    \includegraphics[width=.245\linewidth]{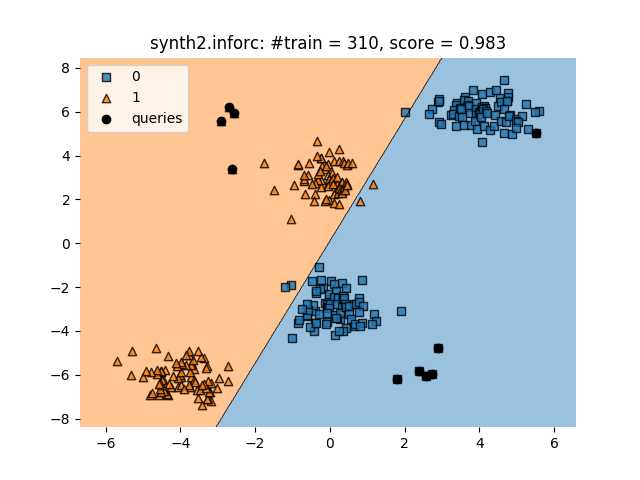}
    \hfill
    \includegraphics[width=.245\linewidth]{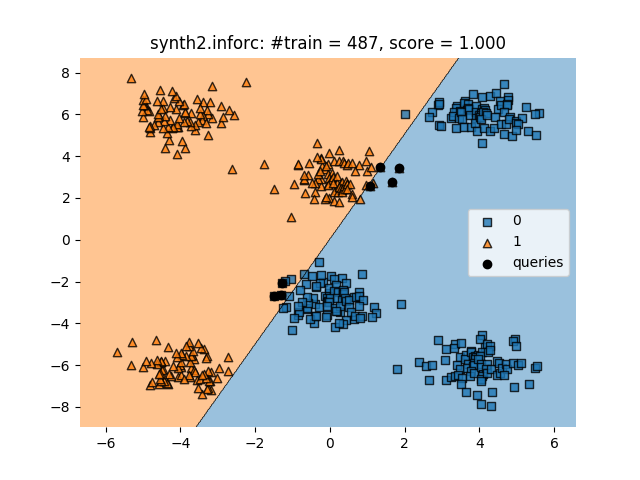}
    \caption{\goral{} under $\goal_{\text{dev}}$ (Oracle)}
  \end{subfigure}

  \begin{subfigure}{\textwidth}
    \centering
    \includegraphics[width=.245\linewidth]{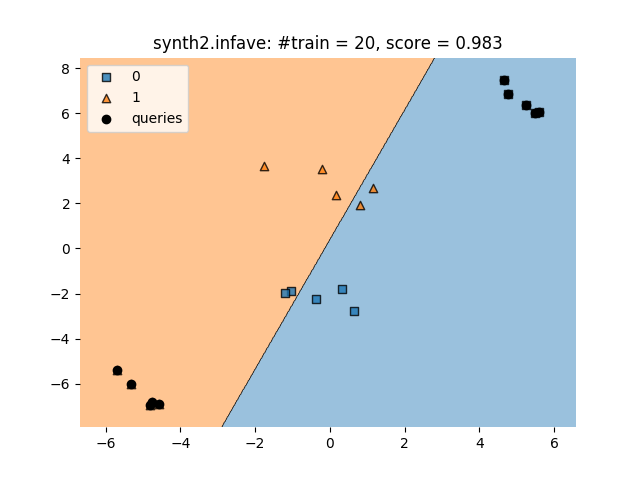}
    \hfill
    \includegraphics[width=.245\linewidth]{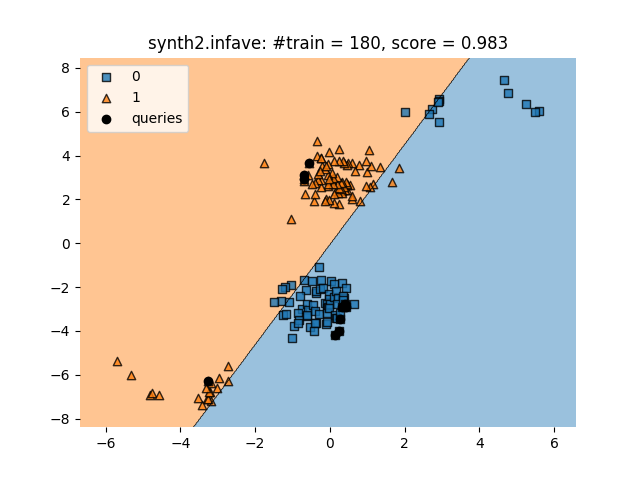}
    \hfill
    \includegraphics[width=.245\linewidth]{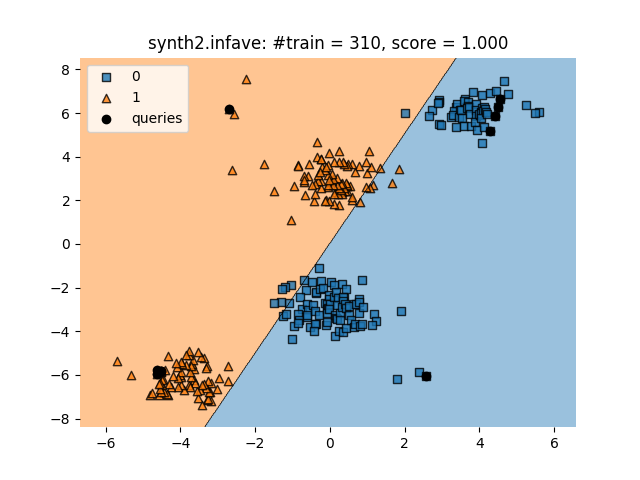}
    \hfill
    \includegraphics[width=.245\linewidth]{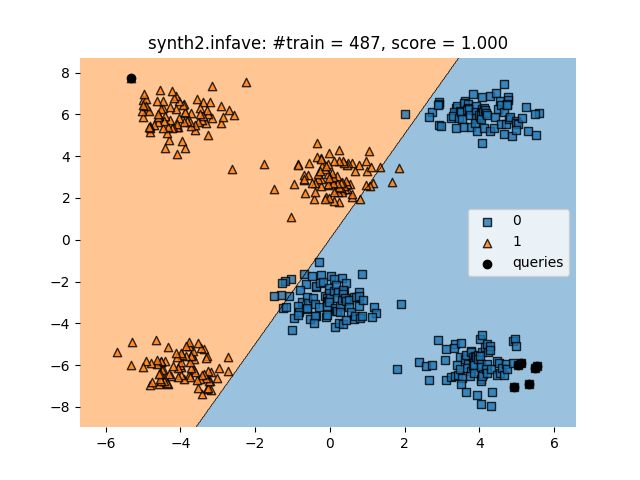}
    \caption{\goral{} under $\goal_{\text{dev}}$ (Average)}
  \end{subfigure}

  \begin{subfigure}{\textwidth}
    \centering
    \includegraphics[width=.245\linewidth]{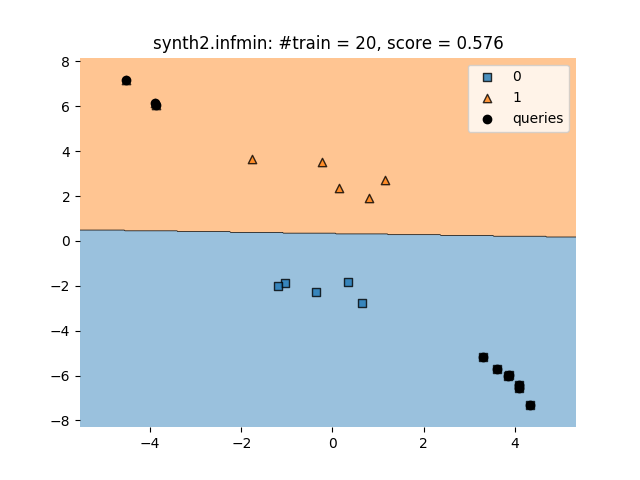}
    \hfill
    \includegraphics[width=.245\linewidth]{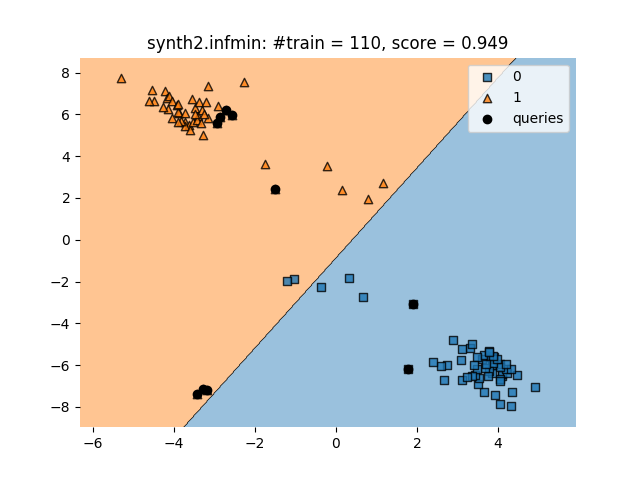}
    \hfill
    \includegraphics[width=.245\linewidth]{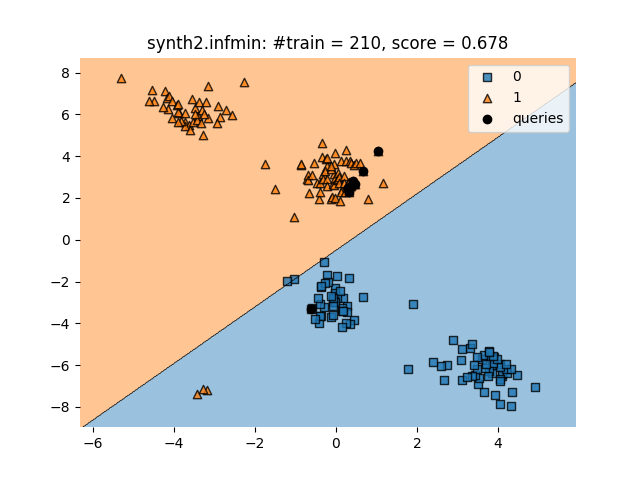}
    \hfill
    \includegraphics[width=.245\linewidth]{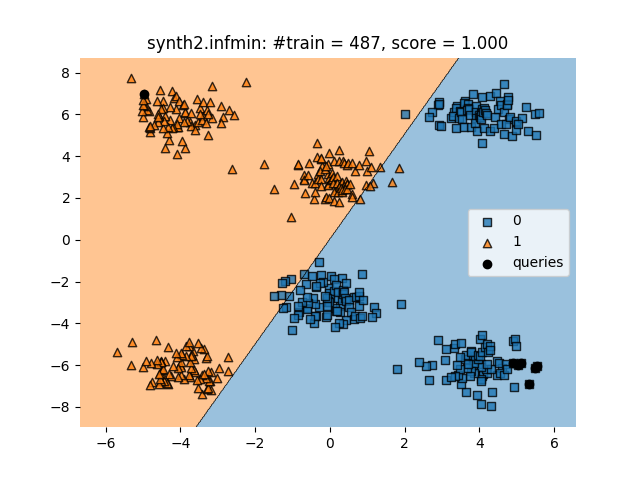}
    \caption{\goral{} under $\goal_{\text{dev}}$ (Minimum)}
  \end{subfigure}
  \caption{AL snapshots on \synth{}}
  \label{fig:synth2_al_iter}
\end{figure}

From the snapshots in Fig.~\ref{fig:synth2_al_iter}, we observe the following querying patterns of the different AL strategies on \synth{}:
\begin{itemize}
  \item Uncertainty sampling: getting stuck with exhausting the two central clusters initially
  \item \goral{} (Oracle): selecting the ``optimal'' samples right away
  \item \goral{} (Average): approximating the Oracle querying pattern quite well, leaving the two distracting clusters to the end
  \item \goral{} (Minimum): starting with the two distracting clusters
\end{itemize}
Also note that even after we take into account the additional $53$ dev-labels exploited by \goral{}, it still enjoys a much higher data efficiency than uncertainty sampling, which reaches the similar level of test accuracy ($0.966$) only after making $230$ queries, $170$ of which are spent (or rather wasted) on the two misleading central clusters.

\subsection{What makes a good goal?}
\label{supp:benchmark}
\begin{figure}[!htb]
  \centering
  \begin{subfigure}{.325\textwidth}
    \centering
    \includegraphics[width=\linewidth]{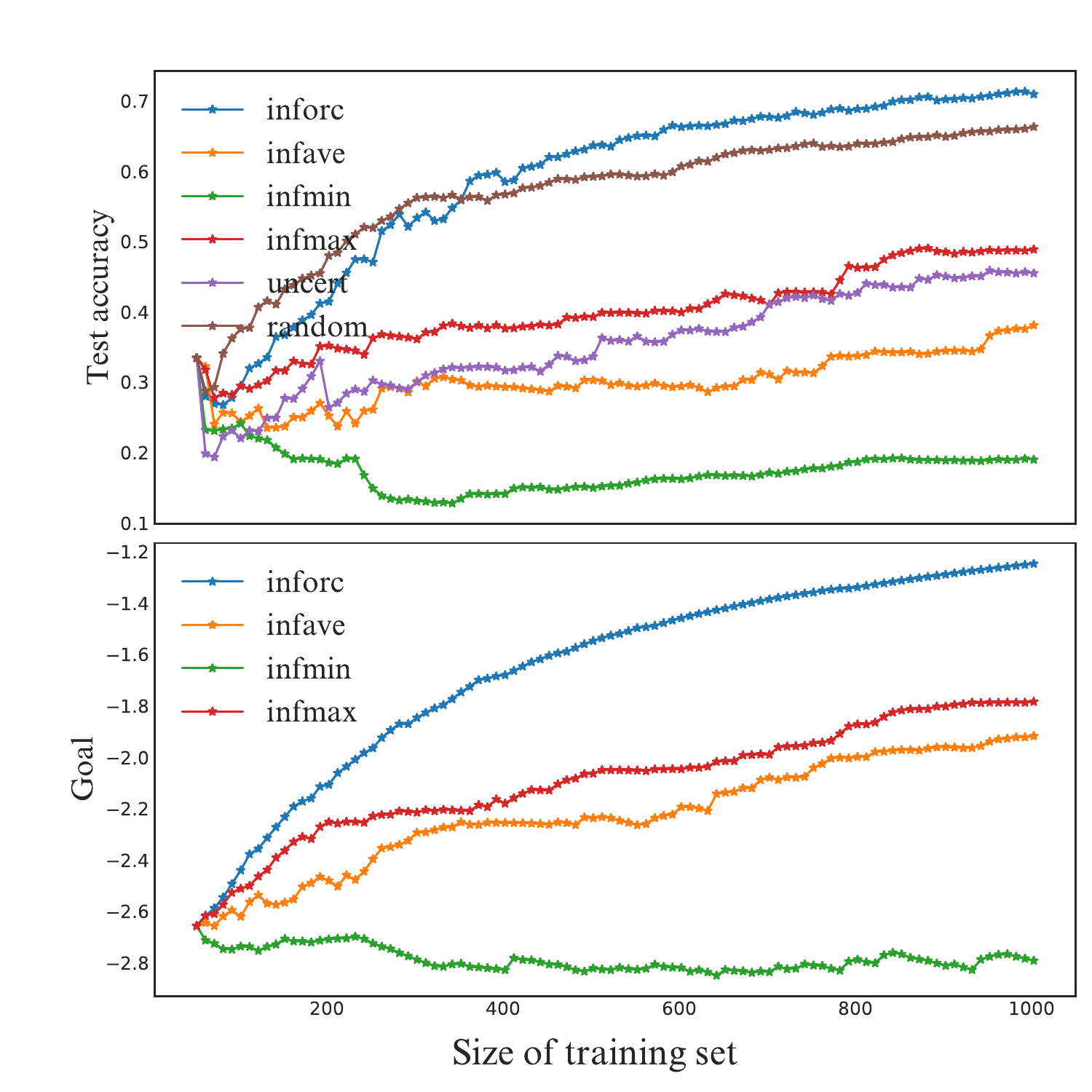}
    \caption{Negative dev-set loss $\goal_{\text{dev}}$}
    \label{fig:letter_dev_al_curve}
  \end{subfigure}
  \hfill
  \begin{subfigure}{.325\textwidth}
    \centering
    \includegraphics[width=\linewidth]{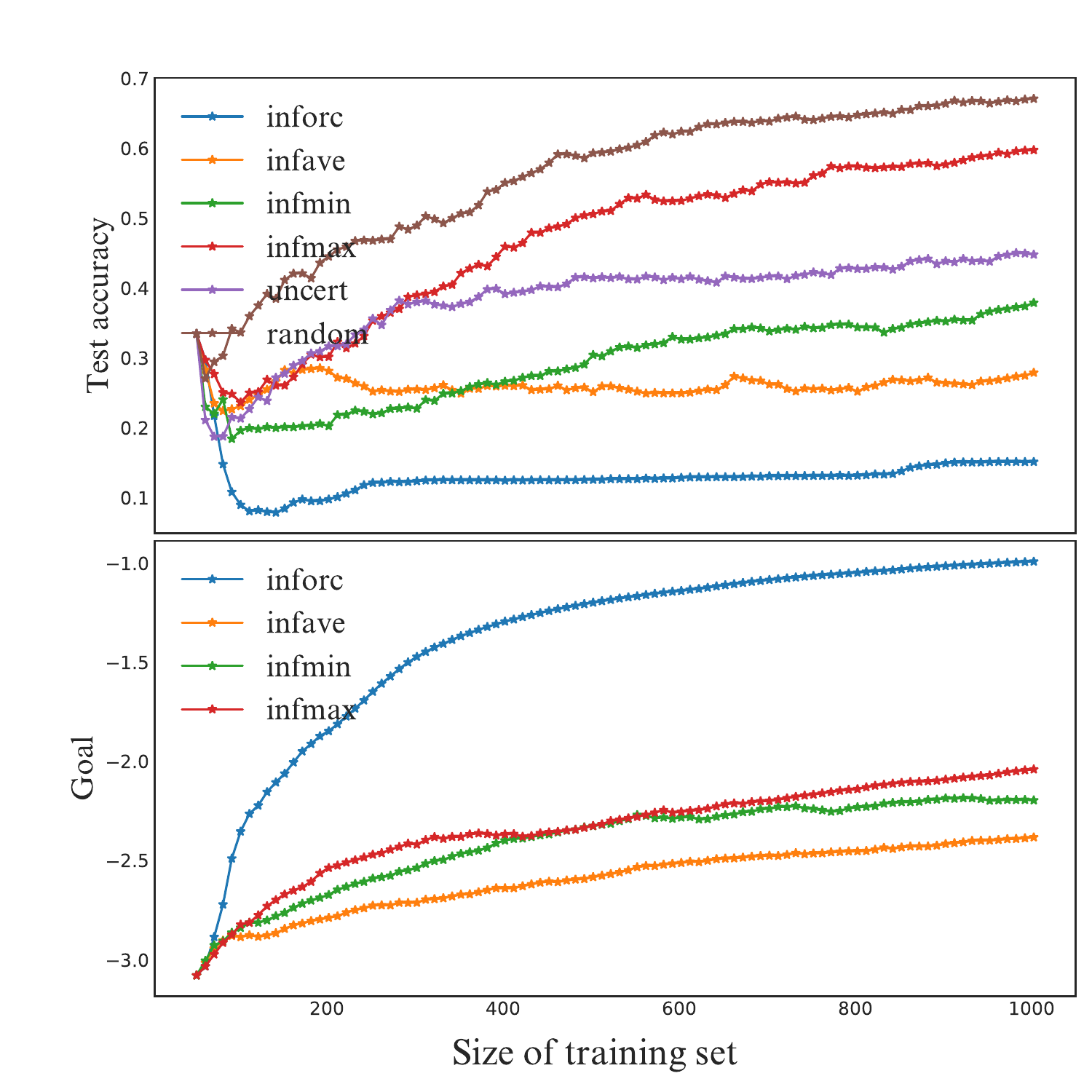}
    \caption{Negative prediction entropy $\goal_{\text{ent}}$}
    \label{fig:letter_ent_al_curve}
  \end{subfigure}
  \hfill
  \begin{subfigure}{.325\textwidth}
    \centering
    \includegraphics[width=\linewidth]{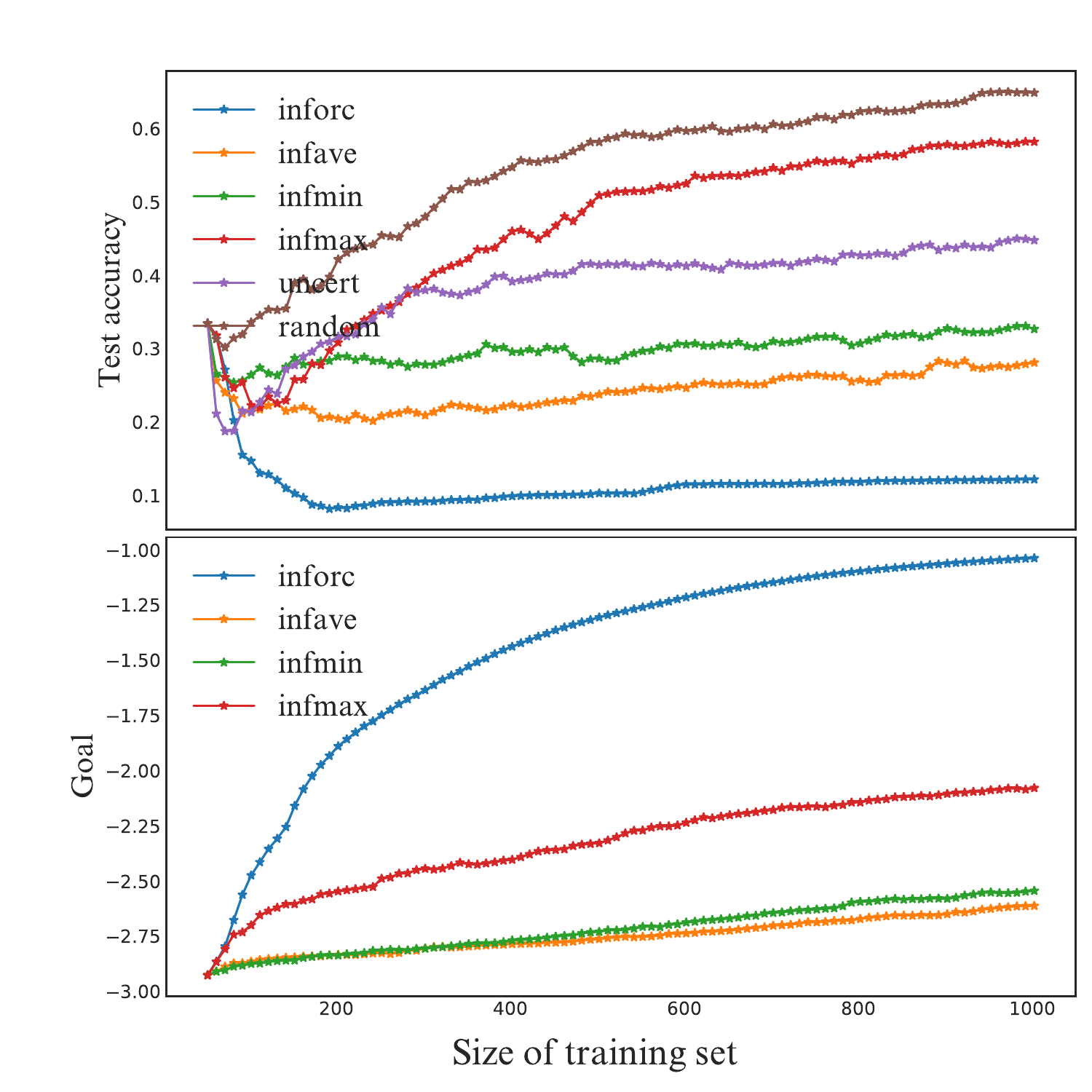}
    \caption{Negative Fisher information $\goal_{\text{fir}}$}
  \end{subfigure}
  \caption{Learning curves and goal curves of various \goral{} strategies on \letter{}.}
  \label{fig:letter_al_curve}
\end{figure}

\begin{figure}[!htb]
  \centering
  \begin{subfigure}{.325\textwidth}
    \centering
    \includegraphics[width=\linewidth]{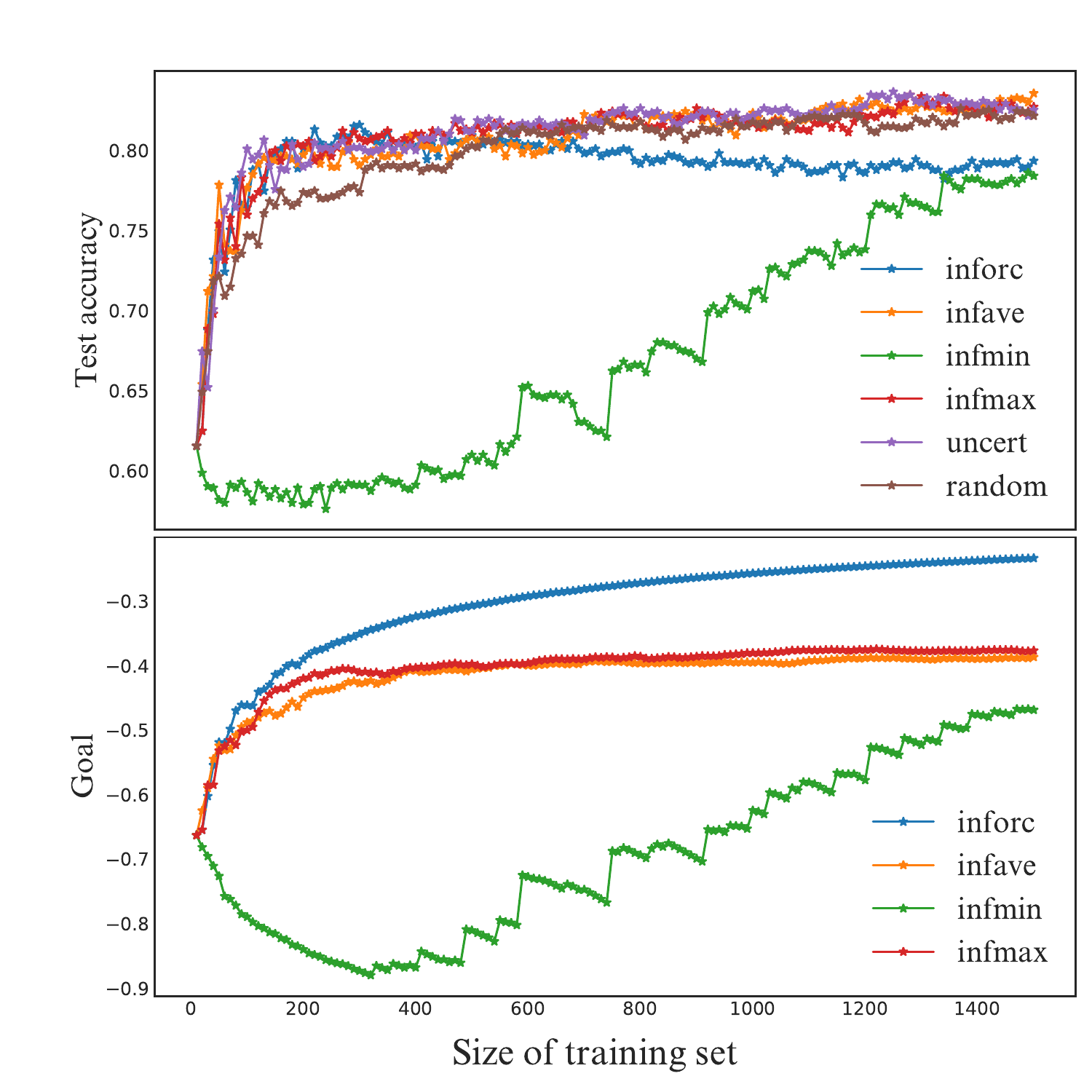}
    \caption{Negative dev-set loss $\goal_{\text{dev}}$}
    \label{fig:rtpolar_dev_al_curve}
  \end{subfigure}
  \hfill
  \begin{subfigure}{.325\textwidth}
    \centering
    \includegraphics[width=\linewidth]{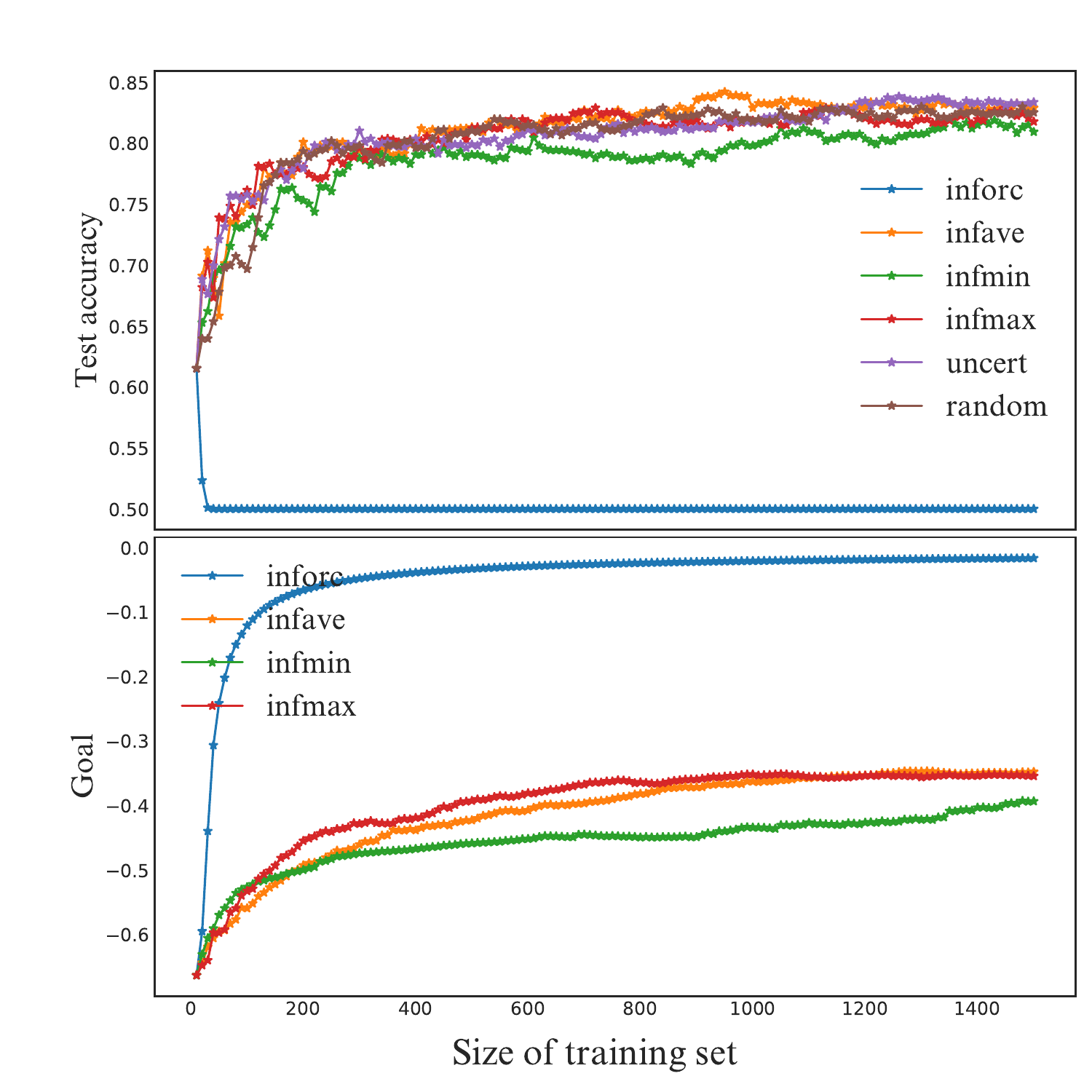}
    \caption{Negative prediction entropy $\goal_{\text{ent}}$}
    \label{fig:rtpolar_ent_al_curve}
  \end{subfigure}
  \hfill
  \begin{subfigure}{.325\textwidth}
    \centering
    \includegraphics[width=\linewidth]{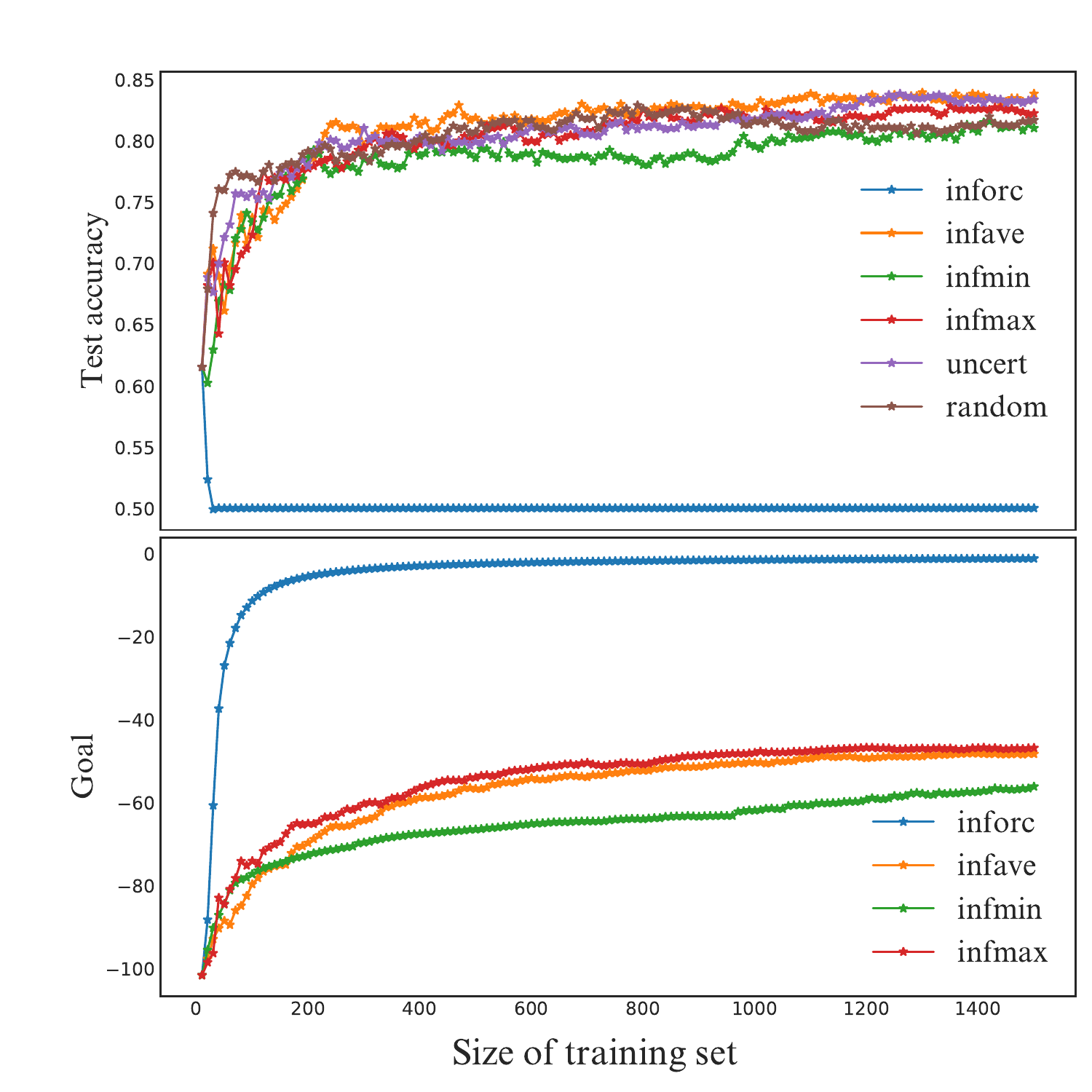}
    \caption{Negative Fisher information $\goal_{\text{fir}}$}
  \end{subfigure}
  \caption{Learning curves and goal curves of various \goral{} strategies on \rtpolar{}.}
  \label{fig:rtpolar_al_curve}
\end{figure}

Below we benchmark \goral{} on two real-world datasets. In Fig.~\ref{fig:letter_al_curve} and~\ref{fig:rtpolar_al_curve}, we show the usual learning curve on the upper half, in which we inspect data efficiency, i.e. how quickly the various strategies help the model reach a certain level of performance (the ``end goal'' in general AL). On the lower half we show the ``goal curve'', in which we look at whether the proposed approximate utility actually helps the model achieving the designated goal (the computable ``proxy goal'' in \goral{}). An ideal \goral{} strategy should find success in both cases.

For $\goal_{\text{dev}}$, we see from Fig.~\ref{fig:letter_dev_al_curve} and~\ref{fig:rtpolar_dev_al_curve} that there is a clear correlation between the two curves, which should not be surprising given the close relationship between dev-set loss and test accuracy. We can also see the over-fitting effect from Fig.~\ref{fig:rtpolar_dev_al_curve} when after about $600$ queries the test accuracy starts to gradually drop despite the still increasing goal (under ``inforc'', i.e. \goral{}-Oracle). Yet Oracle aside, the other practical simple \goral{} strategies do not consistently outperform the baselines.

For $\goal_{\text{ent}}$ and $\goal_{\text{fir}}$, we have seen in Sec.~\ref{Sec:logreg} how they are closely related analytically, and this can also be easily seen from the empirical results below. The most telling message from Fig.~\ref{fig:letter_ent_al_curve} and~\ref{fig:rtpolar_ent_al_curve} is probably the obvious contrast between the two curves (under ``inforc'', i.e. \goral{}-Oracle), where successfully boosting the goal actually leads to the worst AL performance. And this signifies why $\goal_{\text{ent}}$ should not be trusted as a sensible goal in \goral{}. Similar results hold for $\goal_{\text{fir}}$ as well.

\begin{figure}[!htb]
  \centering
  \begin{subfigure}{.325\textwidth}
    \centering
    \includegraphics[width=\linewidth]{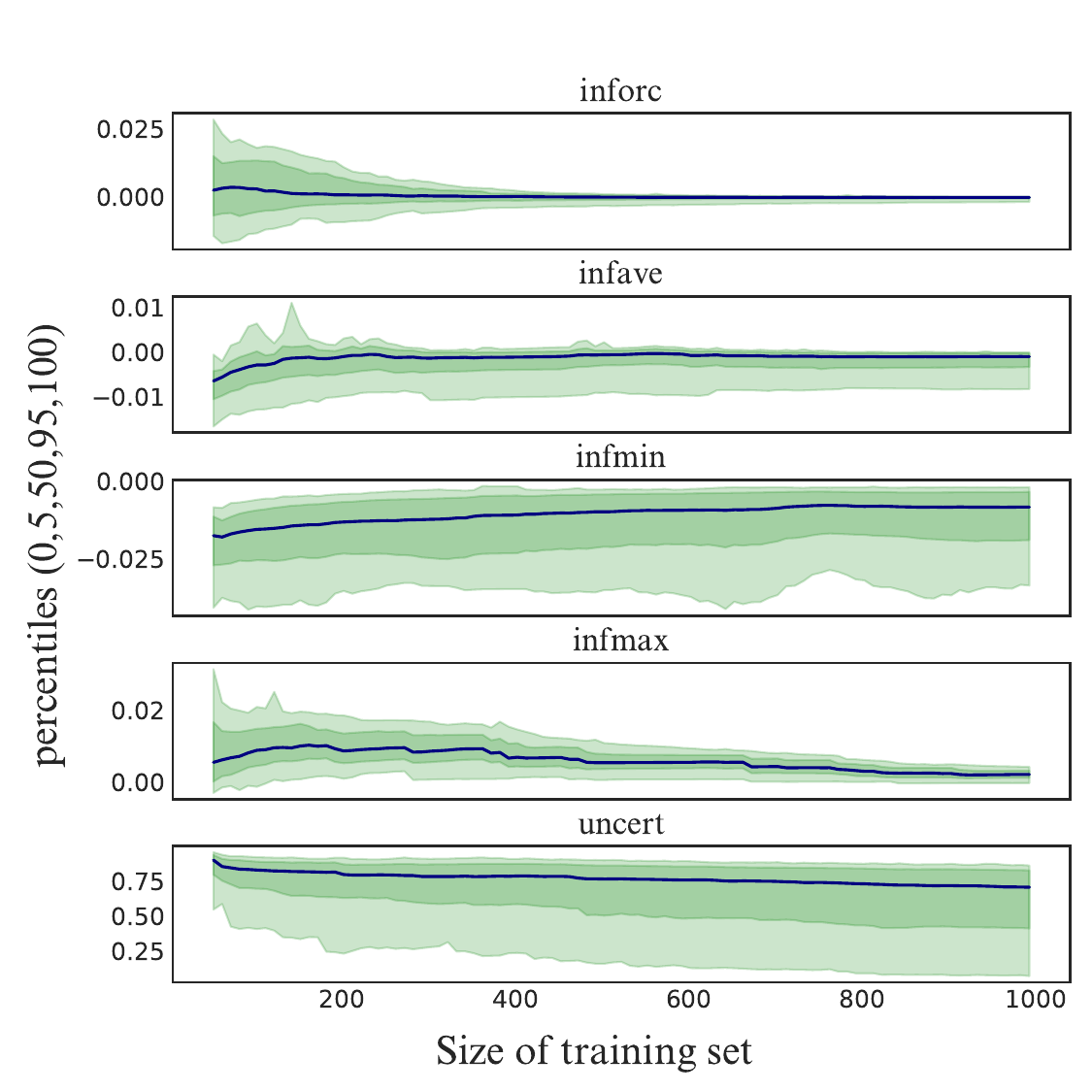}
    \caption{Negative dev-set loss $\goal_{\text{dev}}$}
  \end{subfigure}
  \hfill
  \begin{subfigure}{.325\textwidth}
    \centering
    \includegraphics[width=\linewidth]{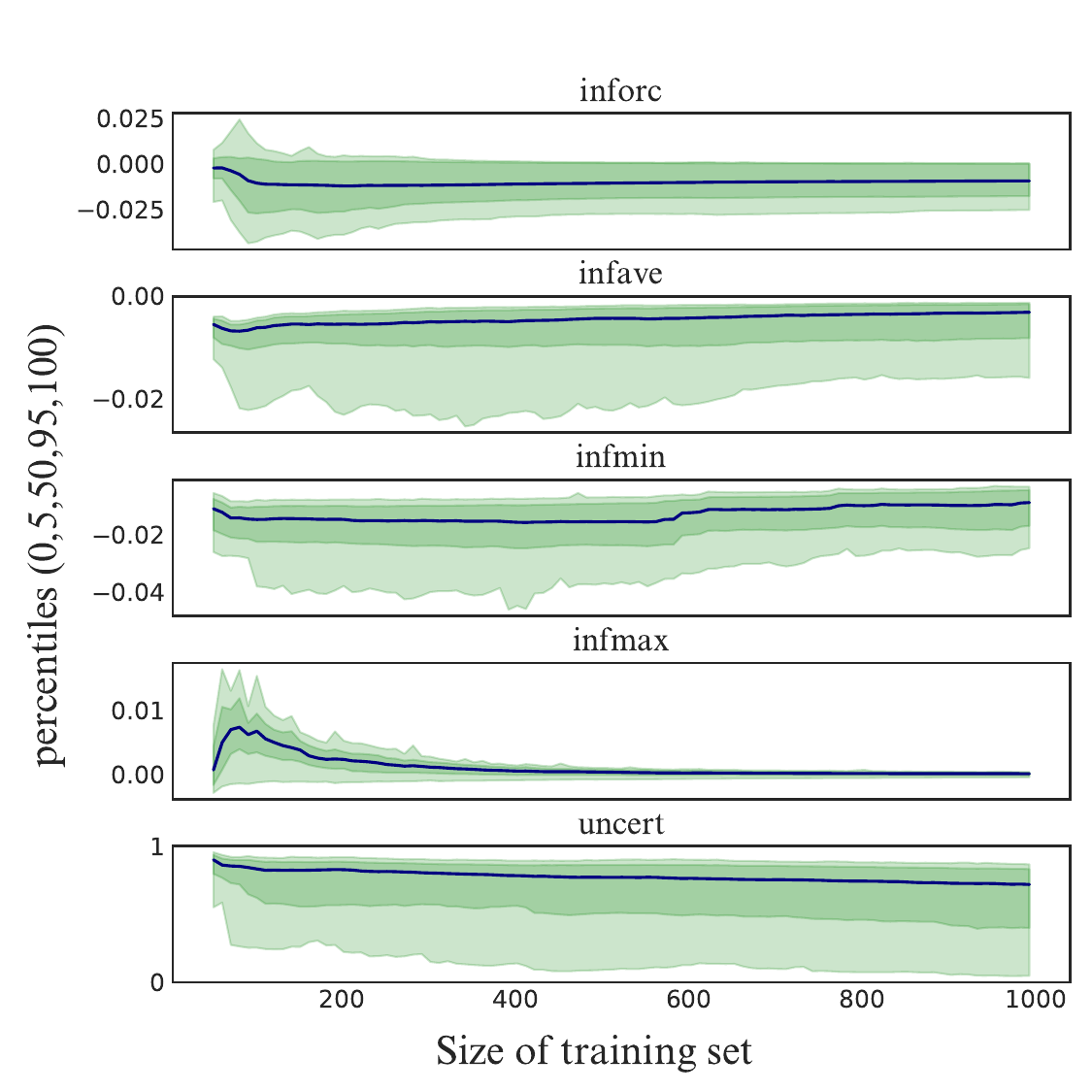}
    \caption{Negative prediction entropy $\goal_{\text{ent}}$}
  \end{subfigure}
  \hfill
  \begin{subfigure}{.325\textwidth}
    \centering
    \includegraphics[width=\linewidth]{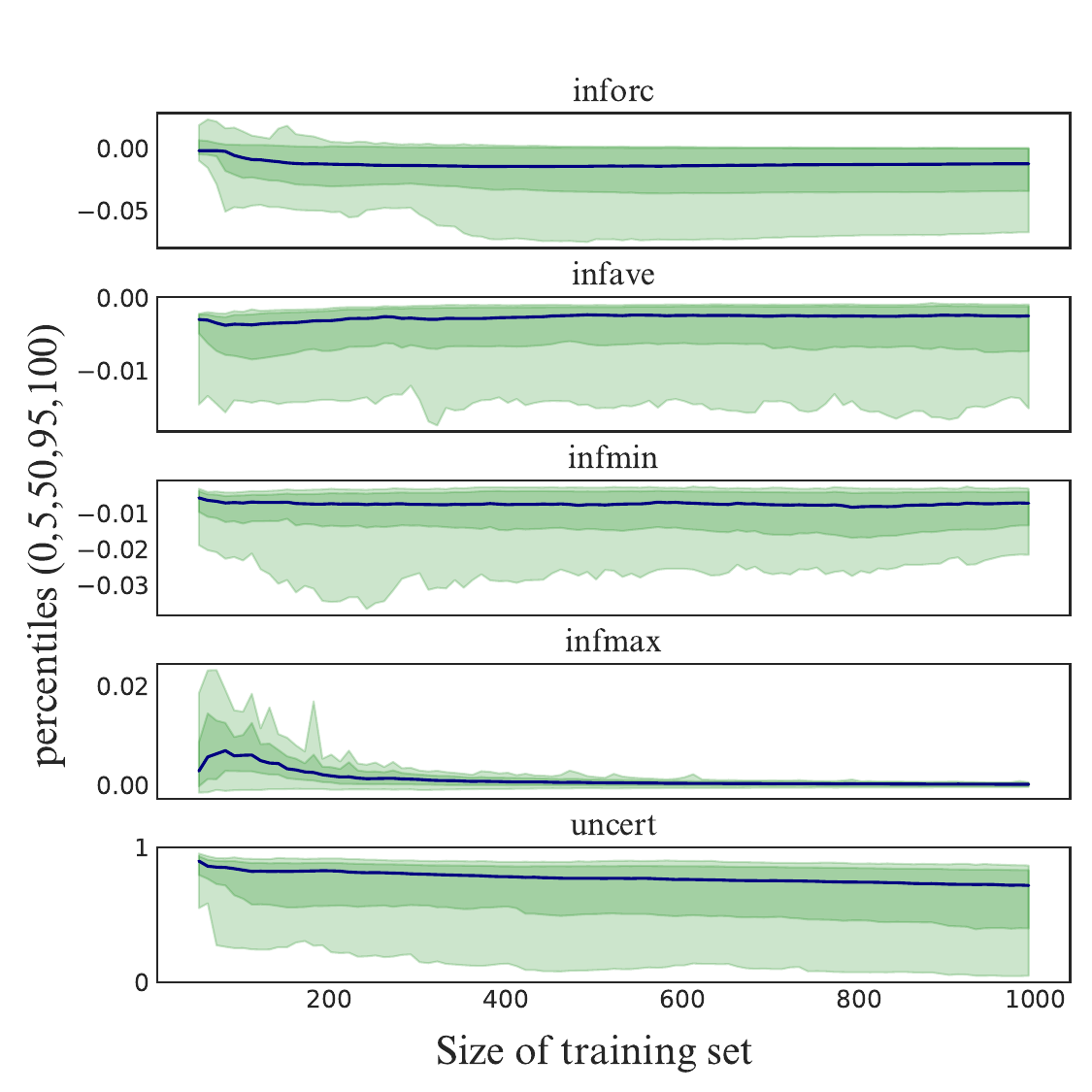}
    \caption{Negative Fisher information $\goal_{\text{fir}}$}
  \end{subfigure}
  \caption{Utility-distribution evolution of various \goral{} strategies on \letter{}.}
  \label{fig:letter_util_distrib}
\end{figure}
\begin{figure}[!htb]
  \centering
  \begin{subfigure}{.325\textwidth}
    \centering
    \includegraphics[width=\linewidth]{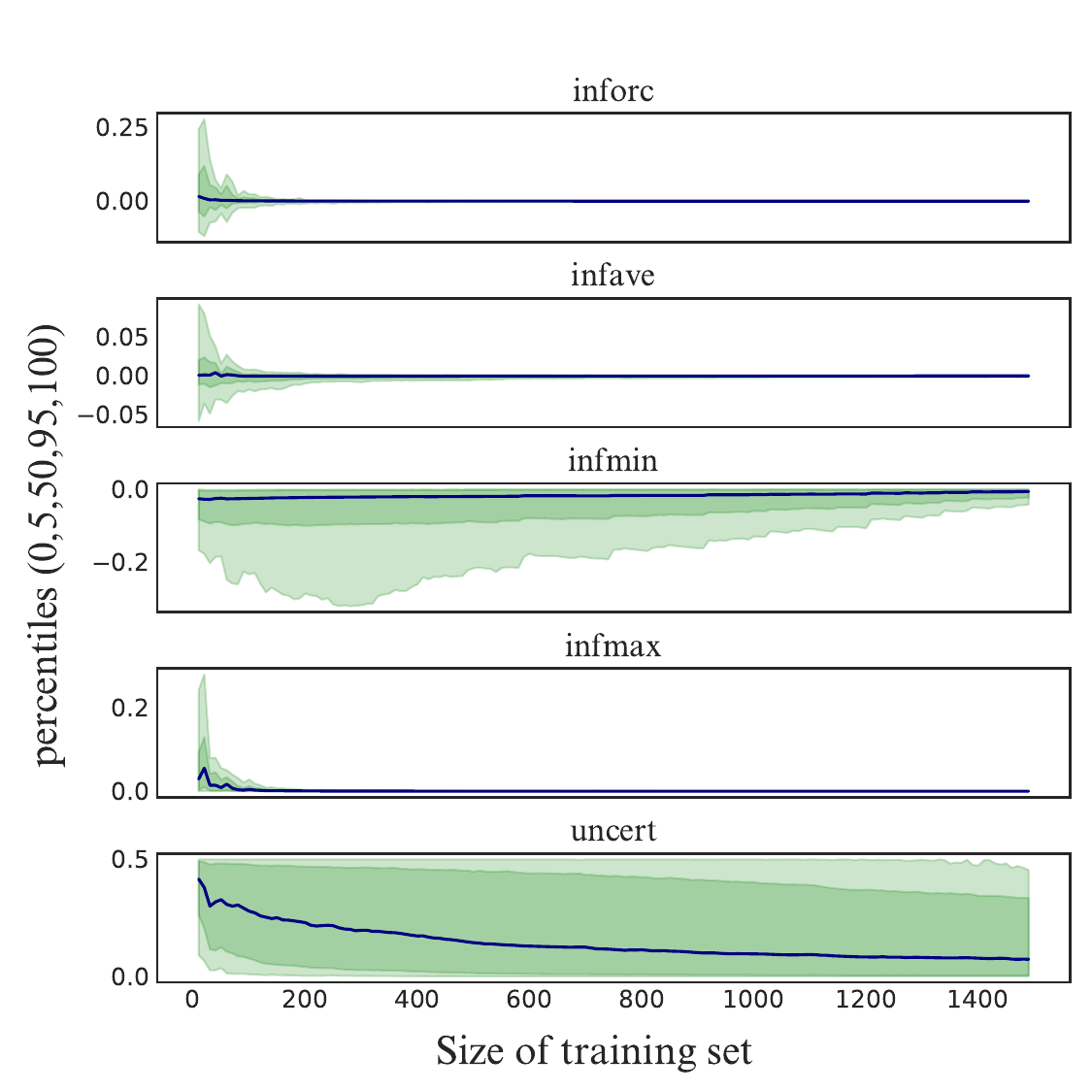}
    \caption{Negative dev-set loss $\goal_{\text{dev}}$}
  \end{subfigure}
  \hfill
  \begin{subfigure}{.325\textwidth}
    \centering
    \includegraphics[width=\linewidth]{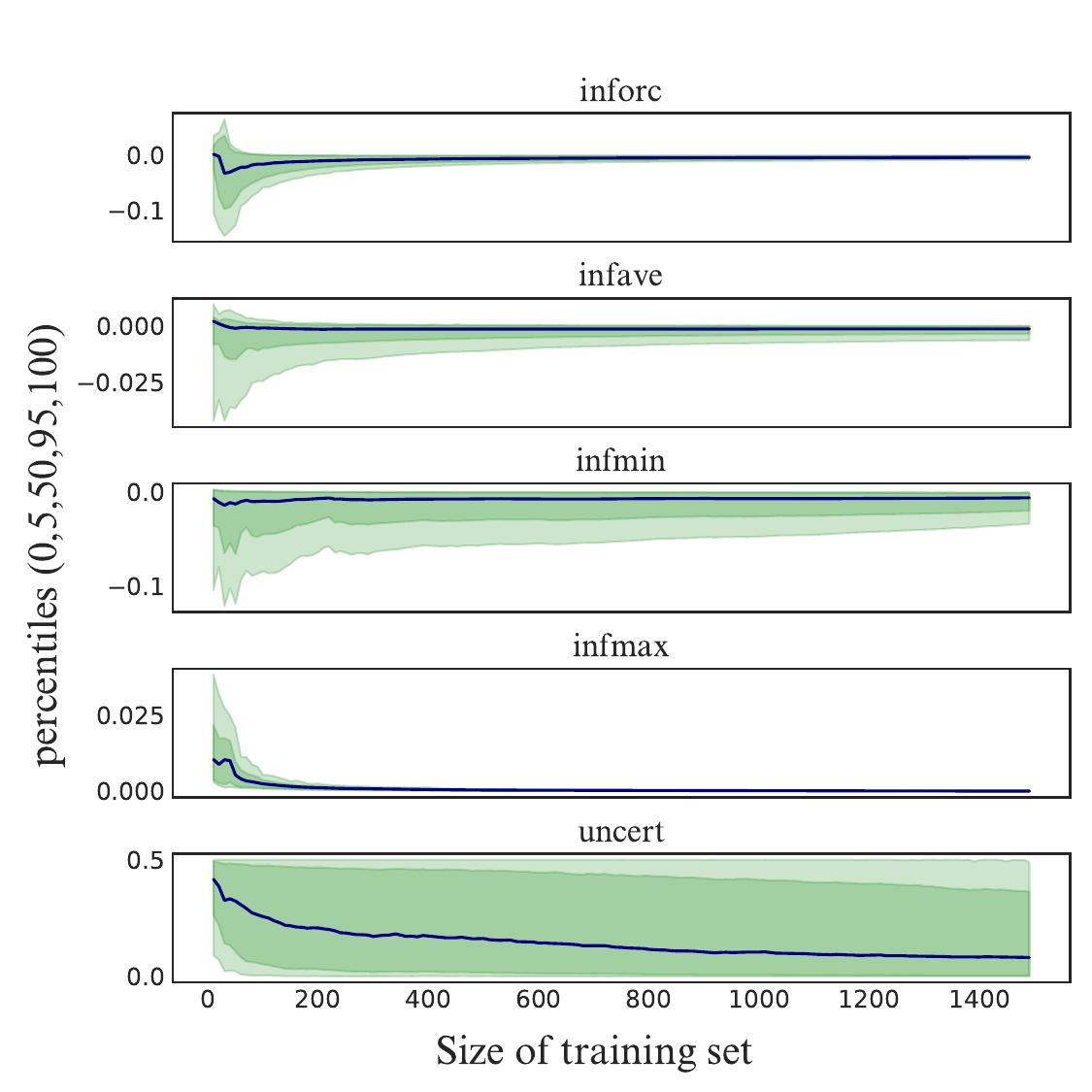}
    \caption{Negative prediction entropy $\goal_{\text{ent}}$}
  \end{subfigure}
  \hfill
  \begin{subfigure}{.325\textwidth}
    \centering
    \includegraphics[width=\linewidth]{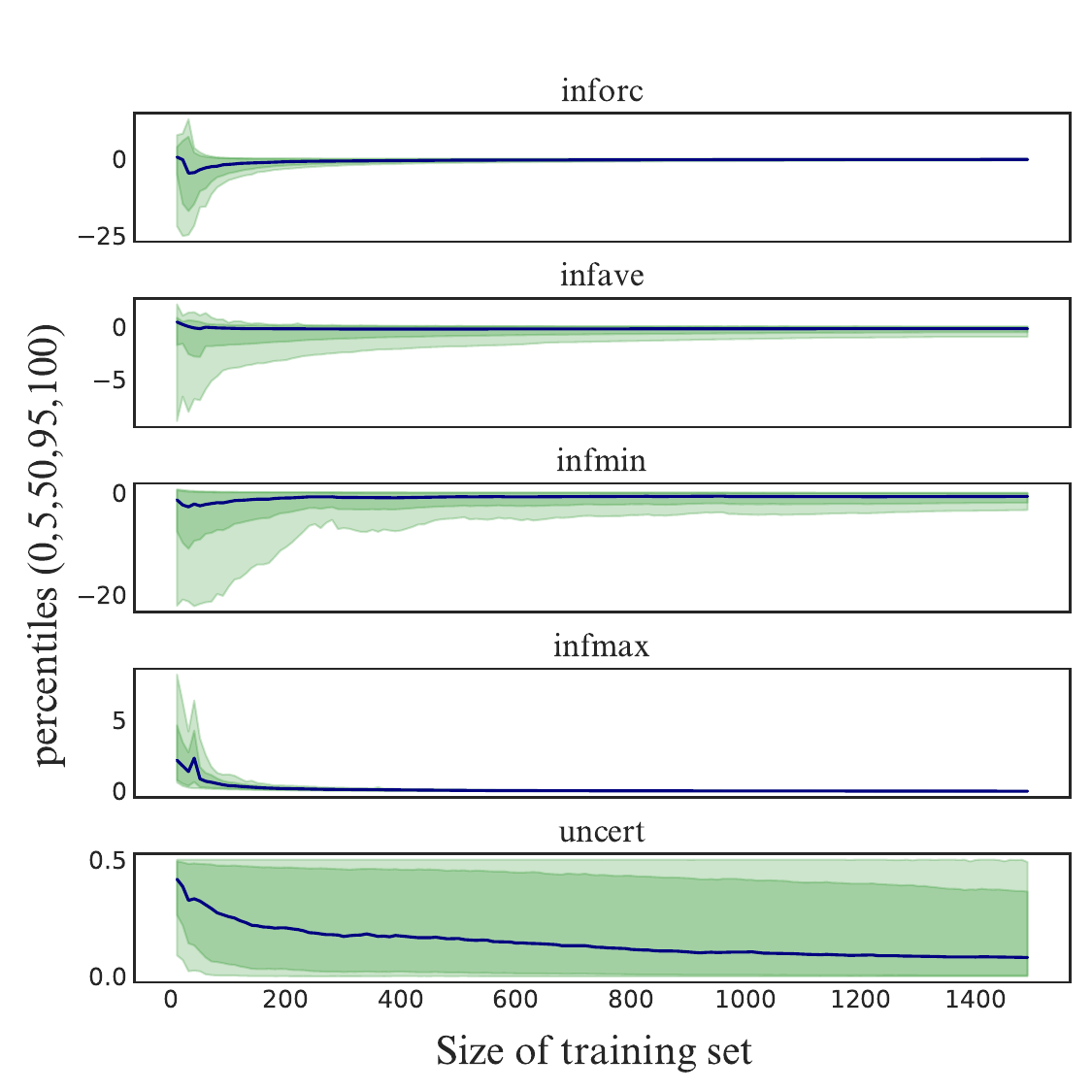}
    \caption{Negative Fisher information $\goal_{\text{fir}}$}
  \end{subfigure}
  \caption{Utility-distribution evolution of various \goral{} strategies on \rtpolar{}.}
  \label{fig:rtpolar_util_distrib}
\end{figure}

\section{A novel goal based on Fisher information}
\label{Supp:goal_fir}
The value of unlabelled data for classification problems in the context of active learning has been studied in~\citep{zhang00pavud}, where Fisher information is used to measure the asymptotic efficiency of parameter estimation and query selection is then aimed at increasing this efficiency. Various notable developments including~\citep{hoi06mic,settles08slt} have been made since then but it was not until fairly recently that~\citep{sourati17fial} first presented a rigorous theoretical investigation into the connection between the popular criterion of ``Fisher information ratio'' used in practice (as well as the various approximations and relaxations therein) and the asymptotic upper bound of the expected variance of the log-likelihood ratio, and thus closed the long-standing gap between theory and practice for works along this line. Below we first briefly recapitulate the main ideas, and then present a novel interpretation of the result which allows us to develop a new goal for the above \goral{} framework.

\paragraph{Fisher information}

Given a parametric probabilistic model $p_{\btheta}(\bx,y)$, Fisher information is defined as the covariance matrix of the score function, i.e. $\fisherinfo(\btheta)\triangleq\expect_{\bx,y}[\nabla_{\btheta}\log p_{\btheta}(\bx,y)\nabla^\top_{\btheta}\log p_{\btheta}(\bx,y)]$. Fisher information can be used to estimate the variance of unbiased parameter estimators (e.g. the maximum-likelihood estimator) due to {\small$\cov[\hat{\btheta}_n]\succeq\fisherinfo(\btheta^{*})^{-1}$} (known as the Cramér–Rao lower bound) and {\small$\cov[\hat{\btheta}_{\infty}]=\fisherinfo(\btheta^{*})^{-1}$}, where $\btheta^{*}$ stands for the ground-truth parameters and $\hat{\btheta}_n$ the parameter estimate from $n$ samples (drawn from $p_{\btheta^{*}}(\bx,y)$).

For discriminative models considered in this paper, parameters $\btheta$ only affect the \emph{conditional} $P(y|\bx)$, i.e. $p_{\btheta}(\bx,y)=p(\bx)P_{\btheta}(y|\bx)$. We therefore additionally define \emph{conditional Fisher information} as $\fisherinfo(\btheta|\bx)\triangleq\expect_{y|\bx}[\nabla_{\btheta}\log P_{\btheta}(y|\bx)\nabla^\top_{\btheta}\log P_{\btheta}(y|\bx)]$. It then naturally follows that $\fisherinfo(\btheta)=\expect_{\bx}[\fisherinfo(\btheta|\bx)]$.

\paragraph{Fisher Information Ratio (FIR)}

Intuitively, one would like to reduce the variance $\cov[\hat{\btheta}_n]$ during learning, yet having a \emph{lower bound} on that is not very helpful. \citep{sourati17fial} show that a different criterion named \emph{Fisher information ratio} actually serves as an asymptotic \emph{upper bound} of the expected variance of the log-likelihood ratio (namely {\small$\log P_{\hat{\btheta}_n}(y|\bx)-\log P_{\btheta^{*}}(y|\bx)$}), i.e.
\begin{align}
\label{Eq:fir}
\expect_{\bx,y}\left[\var_{q}\left(\lim_{n\to\infty}\sqrt{n}\cdot\left[\log P_{\hat{\btheta}_n}(y|\bx)-\log P_{\btheta^{*}}(y|\bx)\right]\right)\right] \leq \trace\left(\fisherinfo_q(\btheta^{*})^{-1}\fisherinfo(\btheta^{*})\right)\text{,}
\end{align}
where $q$ denotes the training distribution $q(\bx,y)=q(\bx)P_{\btheta^{*}}(y|\bx)$, from which training samples (e.g. denoted by $\mathcal{L}_n\triangleq\{(\bx_i,y_i)\}^n_{i=1}$) are drawn that give rise to the estimate $\hat{\btheta}_n$, and $\fisherinfo_q(\btheta^{*})\triangleq\expect_{q(\bx)}[\fisherinfo(\btheta^{*}|\bx)]$. Note that the variance $\var_q(\cdot)$ results from the stochacity of $\mathcal{L}_n$.

Under this criterion, active learning is motivated by selecting queries that form a training distribution $q$ that minimizes FIR, i.e. the r.h.s. of Eq.~\eqref{Eq:fir}, in a hope to quickly reach an estimate $\hat{\btheta}_n$ that has a smaller variance of the log-likelihood ratio. However, solving for the optimal $q$ under FIR is by itself a difficult discrete optimization problem, let alone the fact that computing FIR requires the ground-truth data distribution $p(\bx)$ (for $\fisherinfo(\btheta^{*})$) as well as the true parameters $\btheta^{*}$, neither of which is accessible, and hence calls for further approximation.

\paragraph{Adapting FIR for \goral{}}

First of all, given that $\fisherinfo_q(\btheta^{*})$ and $\fisherinfo(\btheta^{*})$ are both positive semi-definite, we have
$\trace\left(\fisherinfo_q(\btheta^{*})^{-1}\fisherinfo(\btheta^{*})\right) \leq \trace\left(\fisherinfo_q(\btheta^{*})^{-1}\right)\cdot\trace(\fisherinfo(\btheta^{*}))$. Furthermore, in the context of the above \goral{} framework, at every step $q$ effectively represents a discrete distribution supported by the training samples. Hence $\fisherinfo_q(\btheta^{*})=\frac{n}{n+1}\left(\frac{1}{n}\sum_{\bx_i\in\mathcal{L}_n}\fisherinfo(\btheta^{*}|\bx_i)\right) + \frac{1}{n+1}\fisherinfo(\btheta^{*}|\bx')$ (across the various next chosen query $\bx'$) and we therefore approximately treat it as a constant matrix that is independent of $\bx'$, and that leaves us to concentrate on $\trace(\fisherinfo(\btheta^{*}))$ alone. As has been shown in~\citep{chaudhuri15cral,sourati17fial}, under certain regularity conditions, $\fisherinfo(\hat{\btheta}_n)$ provides a fairly good approximation to $\fisherinfo(\btheta^{*})$ (with high probability).

We therefore formulate our FIR-inspired goal to be \textbf{negative Fisher information ($\goal_{\text{fir}}$)} as follows,
\begin{align}
\label{Eq:fir_goal}
\goal_{\text{fir}}(\btheta;\mathcal{U})\triangleq-\trace(\fisherinfo_{u}(\btheta))\text{, where }
\fisherinfo_{u}(\btheta)\triangleq\expect_{\bx\in\mathcal{U}}[\fisherinfo(\btheta|\bx)]=\frac{1}{|\mathcal{U}|}\sum_{\bx\in\mathcal{U}}\fisherinfo(\btheta|\bx)\text{,}
\end{align}
and we effectively use a large pool of unlabelled samples $\mathcal{U}$ to approximate $p(\bx)$.

Below we show that for multinomial logistic regression $\goal_{\text{fir}}$ turns out fairly similar to $\goal_{\text{ent}}$ in nature.

\section{The case for Multinomial Logistic Regression}
\label{Sec:logreg}
Take $\mathcal{X}=\mathbb{R}^d$ and one-hot encoding for the labels, i.e. $\by\in\Delta^{K-1}$ (a $(K-1)$-simplex) that only has one entry (for $y=k$) as 1 and 0 elsewhere. Multinomial logistic regression is parametrized by $\Theta\triangleq(\btheta_1,\dots,\btheta_K)\in\mathbb{R}^{d\times K}$ (or $\btheta=\vectorify(\Theta)\in\mathbb{R}^{dK}$)\footnote{We \emph{over-parametrize} the model by leaving $\btheta_K$ as free parameters rather than fixing it to $\bm{0}_d$, as is normally done in the statistics literature, to make it closer to settings where it is used as the last layer of a neural network.} and encodes $P_{\btheta}(y|\bx)$ through the probability vector $\bp_{\btheta}(\bx)=\sigma(\Theta^\top\bx)\in\Delta^{K-1}$, where $\sigma(\cdot)$ represents the softmax function.

The \emph{per-sample} loss function, as well as its gradient and Hessian matrix, then look as follows,
\begin{align}
R(\bz, \btheta)&=\frac{\lambda}{2}\btheta^\top\btheta - \by^\top\log\bp_{\btheta}(\bx)\text{,}\nonumber\\
\nabla_{\btheta}R(\bz, \btheta)&=\lambda\btheta - \vectorify\left(\bx\cdot(\by - \bp_{\btheta}(\bx))^\top\right)\text{,}\label{Eq:logreg_grad}\\
H(\btheta;\bz)\triangleq\nabla^2_{\btheta}R(\bz, \btheta)&=\lambda I + \Lambda_{\btheta}(\bx)\otimes\bx\bx^\top\text{,}\label{Eq:sample_hessian}
\end{align}
where $\lambda\in\mathbb{R}^{+}$ is the hyper-parameter that controls the strength of $\ell_2$ regularization, $\Lambda_{\btheta}(\bx)\triangleq\diag(\bp_{\btheta}(\bx))-\bp_{\btheta}(\bx)\bp^\top_{\btheta}(\bx)\in\mathbb{R}^{K\times K}$, and $\otimes$ represents the Kronecker product between two matrices. We note that $\Lambda_{\btheta}(\bx)$ is symmetric, diagonally dominant and positive semi-definite, and that the per-sample Hessian $H(\btheta;\bz)$ shown above as well as the full-batch Hessian $H_{\btheta}$ are both symmetric and positive definite, and hence the loss function is convex and $H^{-1}_{\btheta}$ does exist.

\paragraph{Expected utility} As per Sec.~\ref{Sec:if}, the key term in the utility computation is the expected gradient, and
\begin{align}
\expect_y\big[\nabla_{\btheta}R(\bz,\hat{\btheta})\big]&=\nabla_{\btheta}R\big((\bx,\expect_{y}[\by]),\hat{\btheta}\big)\tag*{(due to $\nabla_{\btheta}R(\bz, \btheta)$ being linear to $\by$, per Eq.~\eqref{Eq:logreg_grad})}\\
&=\vectorify\big(\bx\cdot(\bp_{\hat{\btheta}}(\bx) - \bp_y)^\top\big)+\lambda\hat{\btheta}\text{,}\tag*{(per Eq.~\eqref{Eq:logreg_grad} and $\expect_y[\by]=I\cdot\bp_y=\bp_y$)}
\end{align}
where $\bp_y\in\Delta^{K-1}$ represents the probability vector of $P(y)$. From this we also see that when one sets $\bp_y=\bp_{\hat{\btheta}}(\bx)$, $\tilde{\util}_{\text{exp}}(\bx;\hat{\btheta})=\bv^\top_{\hat{\btheta}}\cdot\expect_y\big[\nabla_{\btheta}R(\bz,\hat{\btheta})\big]=\lambda\bv^\top_{\hat{\btheta}}\hat{\btheta}$ and becomes independent of $\bx$.

\paragraph{Gradient of $\goal_{\text{ent}}$}
We derive the gradient for the goal of negative prediction entropy as follows,
\begin{align}
\nabla_{\btheta}\entropy(\bp_{\btheta}(\bx))=-\vectorify\left(\bx\cdot(\bm{1}+\log\bp_{\btheta}(\bx))^\top\Lambda_{\btheta}(\bx)\right)=-\vectorify\left(\bx\cdot(\bp\circ\log\bp+\entropy\bp)^\top\right)\text{,}\nonumber
\end{align}
where we abbreviate $\bp_{\btheta}(\bx)$ and $\entropy(\bp_{\btheta}(\bx))$ with $\bp$ and $\entropy$ respectively in the last step for brevity.

\paragraph{Computing $\goal_{\text{fir}}$}
Below we derive a closed-form solution to the negative Fisher information (Eq.~\eqref{Eq:fir_goal}). We first simplify it by making use of the the well-known result that Fisher information is equal to the expected Hessian of the negative log-likelihood, i.e. $\fisherinfo(\btheta|\bx)=\expect_{y}[H(\btheta;\bz)]=H(\btheta;\bx)$, where the 2nd equation follows from the fact that the Hessian is independent of $y$ as per Eq.~\eqref{Eq:sample_hessian}.

Now we can rewrite the goal as $\goal_{\text{fir}}(\btheta;\mathcal{U})=-\expect_{\bx\in\mathcal{U}}[\trace(H(\btheta;\bx))]$, where
\begin{align}
\trace(H(\btheta;\bx)) &= \trace(\lambda I) + \trace(\Lambda_{\btheta}(\bx))\trace(\bx\bx^\top) = \lambda K + (1-\bp^\top_{\btheta}(\bx)\bp_{\btheta}(\bx))\cdot\bx^\top\bx\text{,}\label{Eq:trace_hessian}\\
\nabla_{\btheta}\trace(H(\btheta;\bx)) &= -2\bx^\top\bx\cdot\vectorify\left(\bx\cdot\bp^\top_{\btheta}(\bx)\Lambda_{\btheta}(\bx)\right)=2\bx^\top\bx\cdot\vectorify\left(\bx\cdot\left((\nu\bm{1}-\bp)\circ\bp\right)^\top\right)\text{,}\nonumber
\end{align}
where we abbreviate $\bp_{\btheta}(\bx)$ with $\bp$ and set $\nu\triangleq\bp^\top\bp$ in the last step. From Eq.~\eqref{Eq:trace_hessian}, we can see that $\goal_{\text{fir}}$, as does $\goal_{\text{ent}}$, also favours models that yield \emph{minimum-entropy} predictions. The close relationship between these two goals is also verified from our empirical studies.

\paragraph{Setting the hyperparameter $\lambda$}

We use the \href{https://scikit-learn.org/stable/modules/generated/sklearn.linear_model.LogisticRegression.html}{Scikit-learn} implementation of MLR in our experiments, which uses a slightly different formulation that involves a regularization constant $C$ and $\lambda=\frac{1}{nC}$ ($n$ being the number of training points). In active learning, $n$ keeps increasing as more labelled samples are added into the training set. To maintain this mapping, we choose to first select $C$ using cross validation\footnote{As a result, we set $C=0.1$ for \synth{} and \rtpolar{} and $C=1$ for \letter{}.} and then update $\lambda$ accordingly during AL iterations.

\end{document}